\documentclass[sigconf,prologue,table]{acmart}

\usepackage{amsmath}
\usepackage{amsfonts}  
\usepackage{bbm} 

\usepackage{graphicx}
\usepackage{textcomp}
\usepackage[table]{xcolor}
\usepackage{xspace}
\usepackage{algorithm}
\usepackage{algpseudocode}
\usepackage[utf8]{inputenc} 
\usepackage[T1]{fontenc}    
\usepackage{hyperref}
\usepackage{booktabs}       
\usepackage{nicefrac}       
\usepackage{microtype}      
\usepackage{microtype}      
\usepackage{tabularx}
\usepackage{wrapfig}
\usepackage{multirow}
\usepackage{epsfig}
\usepackage{subcaption}
\usepackage{marvosym}
\usepackage{color}
\usepackage{threeparttable}
\usepackage{algorithm}
\usepackage{algpseudocode}
\usepackage{setspace}
\usepackage{amsthm}
\usepackage{pifont}
\newcommand{\cmark}{\ding{51}}
\newcommand{\xmark}{\ding{55}}

\usepackage[normalem]{ulem}

\AtBeginDocument{%
  }

\begin{document}

\title{Janus-Q: End-to-End Event-Driven Trading via Hierarchical-Gated Reward Modeling}


\author{Xiang Li}
\email{xli906@connect.hkust-gz.edu.cn}
\authornote{Equal Contribution.}
\affiliation{%
  \institution{The Hong Kong University of Science and Technology (Guangzhou)}
  \city{Guangzhou}
  \country{China}
  \postcode{511458}
}

\author{Zikai Wei}
\email{weizikai@idea.edu.cn}
\authornotemark[1]
\affiliation{%
  \institution{International Digital Economy Academy}
  \city{Shenzhen}
  \country{China}
  \postcode{518000}
}

\author{Yiyan Qi}
\email{qiyiyan@idea.edu.cn}
\affiliation{%
  \institution{International Digital Economy Academy}
  \city{Shenzhen}
  \country{China}
  \postcode{518000}
}

\author{Wanyun Zhou}
\email{wzhou266@connect.hkust-gz.edu.cn}
\affiliation{%
  \institution{The Hong Kong University of Science and Technology (Guangzhou)}
  \city{Guangzhou}
  \country{China}
  \postcode{511458}
}

\author{Xiang Liu}
\email{xliu886@connect.hkust-gz.edu.cn}
\affiliation{%
  \institution{The Hong Kong University of Science and Technology (Guangzhou)}
  \city{Guangzhou}
  \country{China}
  \postcode{511458}
}

\author{Penglei Sun}
\email{psun012@connect.hkust-gz.edu.cn}
\affiliation{%
  \institution{The Hong Kong University of Science and Technology (Guangzhou)}
  \city{Guangzhou}
  \country{China}
  \postcode{511458}
}

\author{Jian Guo}
\email{guojian@idea.edu.cn}
\affiliation{%
  \institution{International Digital Economy Academy}
  \city{Shenzhen}
  \country{China}
  \postcode{518000}
}

\author{Yongqi Zhang}
\email{yongqizhang@hkust-gz.edu.cn}
\affiliation{%
  \institution{The Hong Kong University of Science and Technology (Guangzhou)}
  \city{Guangzhou}
  \country{China}
  \postcode{511458}
}
\authornote{Corresponding author.}

\author{Xiaowen Chu}
\email{xwchu@hkust-gz.edu.cn}
\affiliation{%
  \institution{The Hong Kong University of Science and Technology (Guangzhou)}
  \city{Guangzhou}
  \country{China}
  \postcode{511458}
}
\authornotemark[2]

\renewcommand{\shortauthors}{Xiang Li et al.}

\begin{abstract}
Financial market movements are often driven by discrete financial events conveyed through news, whose impacts are heterogeneous, abrupt, and difficult to capture under purely numerical prediction objectives. 
These limitations have motivated growing interest in using textual information as the primary source of trading signals in learning‑based systems.
Two key challenges hinder existing approaches: (1) the absence of large-scale, event-centric datasets that jointly model news semantics and statistically grounded market reactions, and (2) the misalignment between language model reasoning and financially valid trading behavior under dynamic market conditions.
To address these challenges, we propose \textbf{Janus-Q},
an end-to-end event-driven trading framework that elevates financial news events
from auxiliary signals to primary decision units.
Janus-Q unifies event-centric data construction and model optimization under a two-stage paradigm.
Stage I focuses on event‑centric data construction, building a large‑scale financial news event dataset comprising \textbf{62,400} articles annotated with \textbf{10} fine‑grained event types, associated stocks, sentiment labels, and event‑driven cumulative abnormal return (CAR). Stage II performs decision‑oriented fine‑tuning, combining supervised learning with reinforcement learning guided by a Hierarchical Gated Reward Model (HGRM), which explicitly captures trade‑offs among multiple trading objectives.
Extensive experiments demonstrate that Janus-Q achieves more consistent, interpretable, and profitable trading decisions
than market indices and LLM baselines,
improving the Sharpe Ratio by up to \textbf{102.0\%} while increasing direction accuracy by over \textbf{17.5\%}
compared to the strongest competing strategies.
\end{abstract}




\maketitle

\section{Introduction}

Financial markets have long been studied through the lens of time-series modeling,
where asset prices or returns are predicted directly from historical numerical signals
such as prices, volumes, and technical indicators~\cite{zhou2025deltalag,zhang2025multi}.
Despite decades of methodological advances, purely numerical forecasting remains fundamentally challenging
due to severe noise, non-stationarity, and frequent regime shifts in real-world markets~\cite{fama1970efficient,lo2004adaptive}.
These limitations have motivated growing interest in incorporating alternative information sources,
including financial news, corporate disclosures, and social media, into learning-based trading systems~\cite{zhou2025unleashing,wang2026dynamic,zhou2024finrobot}.

\begin{figure}[t]
    \centering
    \includegraphics[width=1.0\linewidth,
    trim=5pt 0pt 0pt 3pt,
    clip]{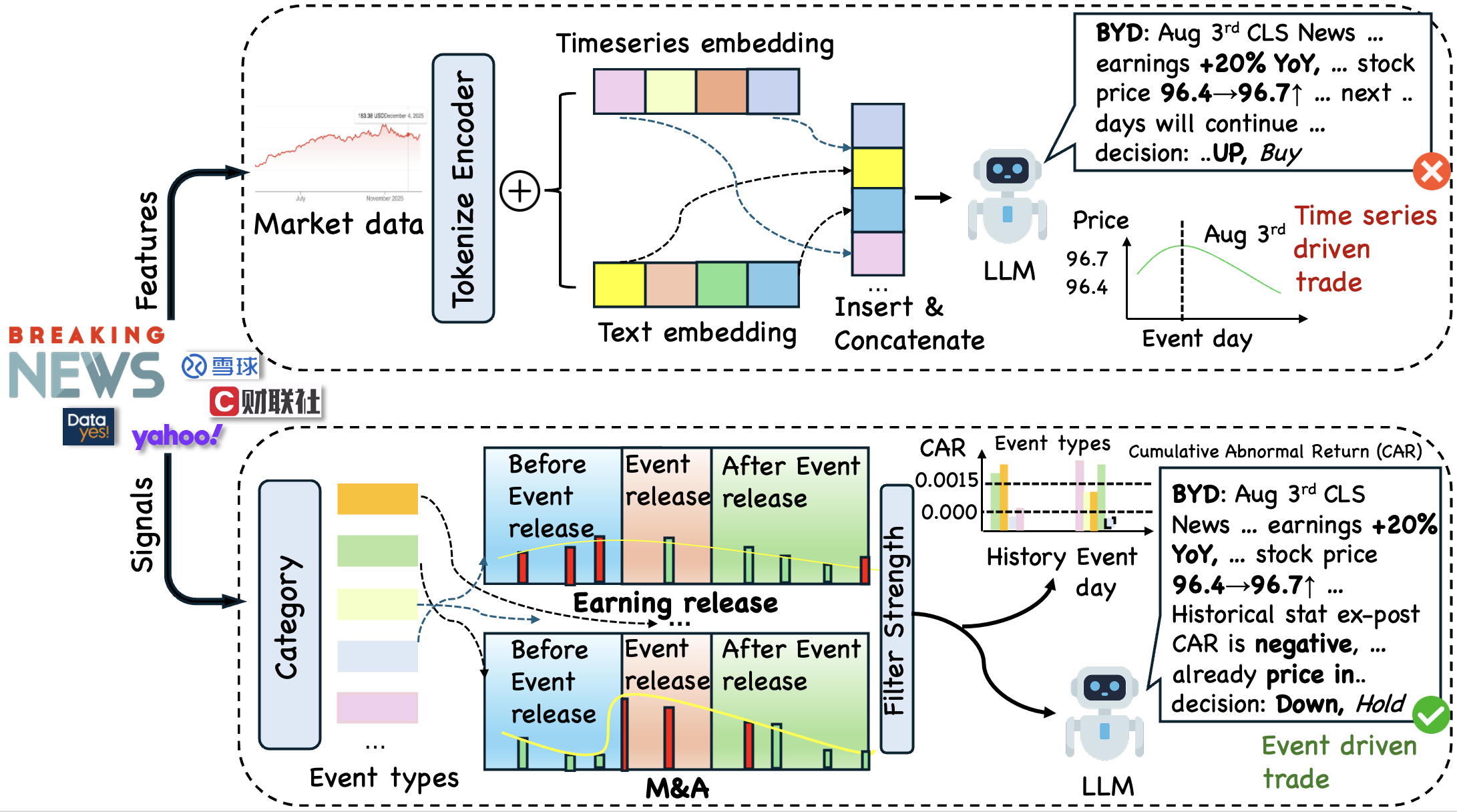}
    \caption{Comparison of traditional time-series–driven trading and event-driven trading strategies.}
    \label{fig:strategy comparison}
\end{figure}

However, a key characteristic of real financial markets is often overlooked:
asset price movements are rarely driven by smooth temporal dynamics alone.
Instead, they are frequently precipitated by discrete and interpretable events,
such as earnings announcements, mergers and acquisitions, or risk disclosures,
which can abruptly shift investor expectations and asset valuations~\cite{mackinlay1997event,thompson1995empirical}.
These events constitute the primary mechanism through which new information is incorporated into prices.
Crucially, different event categories induce highly heterogeneous market responses in terms of direction,
magnitude, and temporal persistence.
Treating such structurally distinct events as homogeneous signals inevitably obscures their economic meaning
and limits decision quality.

Although most market-moving events are communicated through unstructured financial news,
existing learning-based trading systems have yet to fully exploit event-level information.
The dominant paradigm embeds textual news and fuses it with historical price sequences
under numerical forecasting objectives~\cite{kong2025fusing,zhang2025camef}.
Under this formulation, textual information is typically treated as an auxiliary modality,
rather than as a primary driver of trading decisions~\cite{li2024causalstock,dong2024fnspid,saqur2024nifty}.
As a result, learned policies are often dominated by recent price dynamics,
even when those dynamics contradict the semantic implications of newly released events.
Figure~\ref{fig:strategy comparison} illustrates this structural mismatch between time-series–driven
and event-driven trading paradigms.
This limitation stems primarily from the lack of structured supervision at the event level.
While prior work has explored sentiment labels or coarse market reactions~\cite{tetlock2007giving,wang2025pre,li2025finkario},
two fundamental challenges remain largely unaddressed.

\textbf{Challenge 1 (Ch1): Lack of Event–Market Granularity.}
Existing financial datasets fail to jointly model
\emph{what event occurred},
\emph{which assets were affected and with what semantic polarity},
\emph{how the market responded over time},
and \emph{the coverage of event domains}.
Without fine-grained event categories and statistically grounded post-event outcomes,
models struggle to distinguish economically meaningful news from noise
or to reason about heterogeneous market responses across event types~\cite{chen2025dataset,sinha2022sentfin,han2022duee}.
As shown in Table~\ref{tab:comparison}, most existing resources either focus on narrow domains
(e.g., risk or macro events) or provide coarse labels without jointly modeling sentiment and impact magnitude (e.g., CAR).
General-domain datasets suitable for event-driven trading remain particularly scarce.
Moreover, approaches that infer event impact directly from raw price sequences
are heavily confounded by market-wide and firm-specific factors,
making it difficult to isolate causal event effects or learn heterogeneous event-to-market mappings.

\textbf{Challenge 2 (Ch2): Misalignment Between Semantic Reasoning and Market Reality.}
Beyond data limitations, aligning language model reasoning with empirical market behavior remains nontrivial.
Although large language models (LLM) can generate fluent and plausible interpretations of financial news,
their semantic judgments are not inherently grounded in realized market outcomes~\cite{zhang2024multimodal,zhou2025fin}.
In practice, semantic polarity does not map linearly to price movements:
apparently positive announcements may trigger corrections due to priced-in expectations,
while negative news can be absorbed by the market and followed by stabilization or rebound.
Consequently, purely supervised learning risks capturing superficial correlations,
whereas profit-driven optimization alone often induces spurious strategies
that exploit short-term noise~\cite{liu2024dynamic}.

To address these challenges, we propose \textbf{Janus-Q},
a two-stage event-driven trading framework that elevates financial news events
from auxiliary features to primary decision units.
First, to tackle (\textbf{Ch1}), we construct a large-scale financial event dataset comprising
\textbf{62,400} expert-annotated news articles,
each labeled with fine-grained event types, associated stocks, semantic polarity,
and cumulative abnormal returns (CAR), which capture event-induced abnormal
returns by aggregating deviations from expected, event-free price dynamics
over statistically defined event windows.

Building on this dataset, we develop a decision‑oriented training paradigm that directly maps financial news events to executable trading actions.
As illustrated in Figure~\ref{fig:framework}, the framework adopts a multi‑step optimization paradigm.
Supervised fine‑tuning first establishes a reasoning‑aware mapping from event descriptions to expected CAR, effectively integrating textual semantics, market signals, and firm‑specific features.
Subsequently, a Hierarchical Gated Reward Model (HGRM) is introduced for reinforcement fine‑tuning. This component explicitly addresses the semantic–market gap (\textbf{Ch2}) by decomposing the trading reward into interpretable components reflecting event‑type consistency, directional accuracy, and return‑magnitude reliability.
The hierarchical structure further acts as a form of semantic regularization, constraining the learned policy to realize profits through financially grounded reasoning rather than spurious reward exploitation, while maintaining sensitivity to risk and transaction costs.

Extensive experiments demonstrate that Janus-Q consistently outperforms market indices,
time-series–oriented models, and both vanilla and financial LLM baselines.
On average, Janus-Q improves direction accuracy by \textbf{17.5\%}
and event-type accuracy by \textbf{18.2\%} over the strongest competing methods.
Ablation studies further confirm the critical role of HGRM in achieving these gains.
The principal contributions of this paper are as follows:

\begin{itemize}
\item 
We construct a large-scale dataset of \textbf{62,400} financial news articles manually annotated with \textbf{10} event types, associated stocks, semantic labels, and event-driven abnormal returns, forming a unified benchmark for event-level market impact analysis.
\item We propose \textbf{Janus-Q}, the first \underline{end-to-end} event-driven trading framework
that directly maps financial news events to trading decisions,
unifying event interpretation and market response learning via HGRM-guided optimization.

\item 
We evaluate Janus‑Q through extensive experiments and find that it delivers strong and consistent trading performance, improving the Sharpe Ratio by \textbf{102.0\%} and direction accuracy by \textbf{5.4\%} compared with the runner-up strategy, while maintaining a comparable maximum drawdown. These results confirm the effective alignment between event understanding and trading decisions.
\end{itemize}

\begin{table*}[t]
    \centering
    \caption{Comparison of key characteristics of financial event datasets, including event typing, sentiment annotation, market impact magnitude, domain scope, event-type granularity, dataset scale, and language coverage.}
    \label{tab:comparison}
    \renewcommand{\arraystretch}{1.05}
    \setlength{\tabcolsep}{5pt}
    \resizebox{.85\textwidth}{!}{
    \begin{tabular}{l c c c c c c c c}
        \toprule
        \textbf{Dataset} & \textbf{Event Typing} & \textbf{Open Source} & \textbf{Sentiment} & \textbf{Impact Magnitude (CAR)} & \textbf{Domain} & \textbf{\# Event Types} & \textbf{Size} & \textbf{Language} \\
        \midrule
        ChFinAnn~\cite{zheng2019doc2edag} & \cmark & \cmark & \xmark & \xmark & Risky & 5 & 32,040 & CN \\
        DuEE-Fin~\cite{han2022duee} & \cmark & \xmark & \xmark & \xmark & General & 13 & 15,000 & CN \\
        FNSPID~\cite{dong2024fnspid} & \xmark & \cmark & \cmark & \xmark & / & / & 15,700,000 & EN \\
        SEntFiN~\cite{sinha2022sentfin} & \xmark & \cmark & \cmark & \xmark & / & / & 10,753 & EN \\
        CAMEF~\cite{zhang2025camef} & \cmark & \xmark & \cmark & \xmark & Macro & 7 & 2,596 & EN \\
        CSMAR~\cite{wang2025stockmem} & \cmark & \xmark & \xmark & \xmark & General & 13 & / & CN \\
        DocFEE~\cite{chen2025dataset} & \cmark & \cmark & \xmark & \xmark & Risky & 9 & 19,044 & CN \\
        \rowcolor{gray!15}\textbf{Ours} & \cmark & \cmark & \cmark & \cmark & General & 10 & 62,400 & CN \\
        \bottomrule
    \end{tabular}}
\end{table*}

\section{Related Work}

\subsection{Event-Driven Modeling}

Early studies on event-driven market analysis originate from econometric and statistical finance, where discrete corporate or macroeconomic events are treated as exogenous shocks and analyzed through abnormal return dynamics around predefined event windows. The classical event study methodology formalized this process by estimating expected returns from factor models and attributing deviations to the information content of events, establishing a standard toolkit for measuring market reactions \cite{mckinlay1997event}. 
As textual disclosures and news became increasingly accessible, subsequent research incorporated qualitative information into event analysis, showing that media tone and sentiment convey economically meaningful signals that explain and predict short-horizon market movements beyond purely price-based features \cite{tetlock2007}.
Building on this line, machine learning (ML) and deep learning (DL) approaches began to integrate textual information with market data, shifting from handcrafted sentiment indicators toward learning event semantics directly from text. 
Ding et al.~\cite{ding2015event} extract structured events from news and learn dense event representations to model both short-term and long-term effects on stock movements, with subsequent work further enhancing event representations by incorporating richer contextual and temporal signals \cite{ding2016event}. 
More recently,
Wang et al.~\cite{wang2025stockmem} introduce StockMem, an event-reflection
memory framework that organizes news into structured events and leverages
their temporal evolution to retrieve analogous historical scenarios,
supporting more explainable stock movement prediction.
However, despite richer event representations, existing pipelines remain prediction‑oriented and treat events primarily as inputs, without explicitly modeling event‑level properties such as historical context or relative importance to support decision‑oriented trading objectives. Moreover, the field still lacks a comprehensive, standardized dataset for trading. Following this direction, we propose an event‑centric dataset that integrates event categorization, sentiment annotation, and CAR‑based evaluation of event impacts.

\subsection{LLM for Trading}

Recent studies have explored LLMs for trading-related tasks, motivated by their ability to process unstructured financial texts and to provide explicit reasoning through chain-of-thought generation~\cite{xie2024finben,yu2024fincon}. Unlike classical ML or DL models that operate as black-box predictors relying on past numerical data and struggling with abrupt market shifts, LLMs can articulate intermediate causal rationales when interpreting complex market events and respond adaptively to sudden market shocks~\cite{tatsat2025beyond,cao2025deep}.
Yang et al.\cite{yang2025fingptopensourcefinanciallarge} adapt instruction‑tuned LLM for financial analysis and trading, showing that they effectively extract market‑relevant semantics from financial texts. Zhang et al.\cite{zhang2024multimodal} propose \textit{FinAgent}, an LLM‑based financial agent that interacts with market environments and external tools to support trading and portfolio management.
Xiao et al.~\cite{xiao2024tradingagents} study multi-agent trading frameworks powered by LLM, where language models coordinate decision-making through communication and tool usage in simulated markets. More recently, Xiao et al.~\cite{xiao2025trading} introduce \textit{Trading-R1}, which applies reinforcement learning to enhance LLM reasoning for trading decisions, highlighting the potential of RL-based fine-tuning for aligning language model outputs with trading actions. In parallel, Li et al.~\cite{lin2025retuning} propose \textit{RETuning}, a framework that improves stock movement prediction by refining inference-time reasoning with reflective and evidence-based analysis over rich textual inputs.
Despite these advances, current LLM-based trading systems face two complementary limitations. First, trading decisions are often treated as opaque actions primarily driven by price or volume signals, with textual information only weakly integrated into the decision process. Second, reinforcement learning–based methods typically rely on heuristic, linearly additive reward designs in which competing objectives may offset one another, limiting their ability to model economically meaningful trade-offs. Motivated by these limitations, we propose a \textit{Hierarchical-Gated Reward Model} that aligns event-level semantic reasoning with market-grounded trading outcomes.

\begin{figure*}[ht]
  \centering
  \includegraphics[
    width=0.82\textwidth,
    trim=5pt 20pt 5pt 20pt,
    clip
  ]{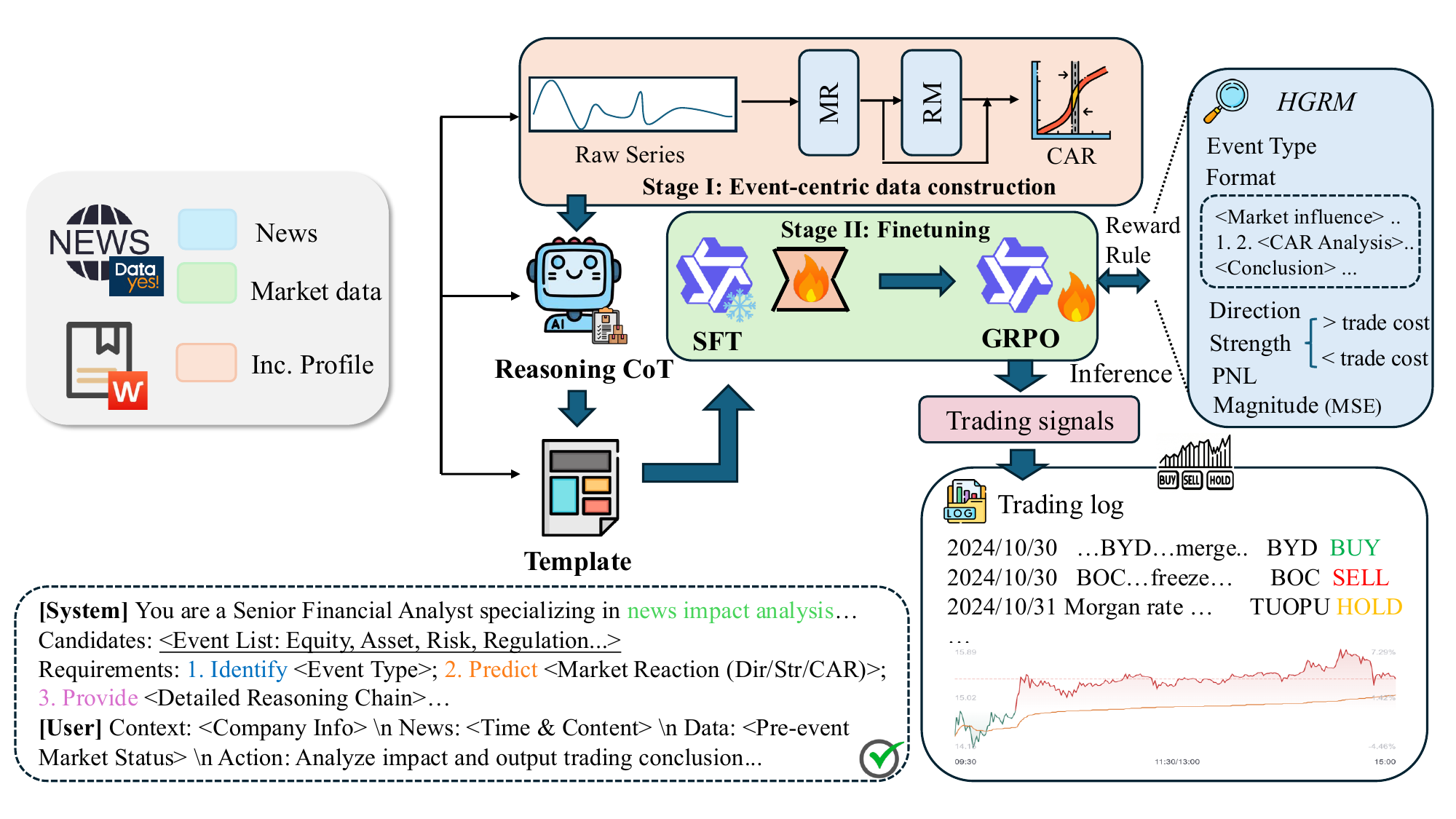}
  \caption{Overview of the proposed two-stage event-driven trading framework.
  Stage~I focuses on event-centric data construction with market-grounded supervision,
  while Stage~II performs decision-oriented fine-tuning via supervised and reinforcement learning.}
  \label{fig:framework}
\end{figure*}

\section{Methodology}

\label{sec:method}

This section describes the \textbf{why} and \textbf{how} of event-driven trading.
As illustrated in Figure~\ref{fig:framework}, our framework is divided into two core paradigms:
Stage~I, which focuses on data construction, and Stage~II, which focuses on model training.

\subsection{Event Centric Data Construction}
\label{sec:event_to_car}
\textbf{Why}: The natural question guiding the collection of this dataset is:
\emph{how will this event affect the associated asset in terms of direction and magnitude,
and how does its impact differ from that of other event types?
} \textbf{How}: To address this, 
we construct an event-centric dataset that captures both the market impact
and the semantic interpretation of financial news events.
Specifically, motivated by classical event study methodology~\cite{mackinlay1997event},
we first quantify impact magnitude through Event-to-CAR modeling,
and then associate each event with category labels and structured semantic annotations
to support event-level supervision. Event type annotations are conducted by a panel of six domain professionals, including fund researchers and securities analysts, ensuring consistency,
financial relevance, and reliability.


\subsubsection{Event-to-CAR Modeling}


Given a news event for stock $i$ occurring at event time $t_0$,
we quantify its market impact using the CAR
computed over a predefined event window.
Let $r_{i,t}$ denote the return of stock $i$ on trading day $t$.
We define two time intervals: an \emph{estimation window}
$\mathcal{T}_{\mathrm{est}} = (T_0,\, T_1]$
precedes the event and is used to estimate event-free normal returns, and 
an \emph{event window}
$\mathcal{T}_{\mathrm{evt}} = (T_1,\, T_2]$
captures abnormal price reactions associated with the event,
including potential pre-event leakage and delayed market adjustment.
Here $T_0, T_1,$ and $T_2$ specify the temporal boundaries of the estimation
and event windows relative to $t_0$.
The timing structure is illustrated in Figure~\ref{fig:timeline}.



\noindent $\bullet$ \textbf{Market Model (MR)}
To remove broad market movements, we estimate the expected return of each stock $i$
using the market model over the estimation window $\mathcal{T}_{\mathrm{est}}$:
\begin{equation}
r_{i,t} = \alpha_i + \beta_i r_{m(i),t} + \varepsilon_{i,t},
\quad t\in\mathcal{T}_{\mathrm{est}}.
\label{eq:market_model}
\end{equation}

where $r_{m(i),t}$ is the return of the benchmark market index selected for stock $i$.
We choose the benchmark index $m(i)$ according to the market-cap segment of stock $i$
(e.g., large-/mid-/small-cap) so that $r_{m(i),t}$ matches its investable universe.


Given the parameters $\hat{\alpha}_i$ and $\hat{\beta}_i$ estimated from the
estimation window $\mathcal{T}_{\mathrm{est}}$ via an ordinary least squares fit
of the market model \cite{kolari2010event}, we evaluate the fitted model on the event window
to obtain abnormal returns (AR) as
\begin{equation}
AR^{\mathrm{MR}}_{i,t} = r_{i,t} - \bigl(\hat{\alpha}_i + \hat{\beta}_i r_{m(i),t}\bigr),
\quad t\in\mathcal{T}_{\mathrm{evt}}.
\label{eq:ar_mm}
\end{equation}

\noindent $\bullet$ \textbf{Risk Model (RM) Neutralization}
Market-adjusted returns may still contain systematic style/industry effects.
We further neutralize $AR^{\mathrm{MR}}_{i,t}$ using a multi-factor risk model on $\mathcal{T}_{\mathrm{evt}}$:
\begin{equation}
AR^{\mathrm{MR}}_{i,t}
= \gamma_i + \mathbf{x}_{i,t}^{\top}\boldsymbol{\lambda}_t + u_{i,t},
\quad t\in\mathcal{T}_{\mathrm{evt}}.
\label{eq:barra}
\end{equation}
where $\mathbf{x}_{i,t}$ denotes the Barra factor exposures of stock $i$ (e.g., style and industry exposures under risk model),
$\boldsymbol{\lambda}_t$ are the corresponding factor returns/premia at time $t$,
$\gamma_i$ is a stock-specific intercept capturing persistent effects not explained by the factors,
and $u_{i,t}$ is the idiosyncratic component.

We apply the estimated factor premia to obtain the \emph{factor-neutral}
abnormal return within the event window:
\begin{equation}
AR_{i,t}
= AR^{\mathrm{MR}}_{i,t}
- \mathbf{x}_{i,t}^{\top}\widehat{\boldsymbol{\lambda}}_t,
\quad t \in \mathcal{T}_{\mathrm{evt}}.
\label{eq:ar_final}
\end{equation}
Details of the risk model specification and factor definitions
are provided in Appendix~\ref{app:risk_model}.

\noindent\textbf{Cumulative Abnormal Return (CAR).}
We then aggregate abnormal returns over the event window to obtain
the cumulative abnormal return:
\begin{equation}
CAR_i
= \sum_{t \in \mathcal{T}_{\mathrm{evt}}} AR_{i,t}.
\label{eq:car}
\end{equation}
The resulting CAR summarizes the total abnormal price movement
attributable to the event, capturing both potential pre-event information
leakage and delayed post-event market adjustment.



\subsubsection{Event Taxonomy and Sentiment Annotation}

The cumulative abnormal return (CAR) defined above serves as the \emph{impact magnitude}
of an event, quantifying its realized economic effect on asset prices.
To complement this market-grounded signal, we further associate each event with
structured semantic labels.

Formally, each event is represented by a ground-truth label
$y = \{c, d, s, e\}$,
where $c$ denotes the realized CAR and $e \in \mathcal{E}$ is the annotated event type.
In our formulation, the direction $d$ (positive, negative, or neutral)
and the trading strength $s$ (strong or weak)
are deterministically derived from $c$:
the direction is obtained via the sign function $d = \mathrm{sign}(c)$,
while the trading strength is determined by a predefined threshold $\tau$
that specifies whether the magnitude of the market impact is sufficient
to justify the execution of the trade.
Together, $d$ and $s$ serve as \emph{semantic} that characterizes
how the market reacts to an event, complementing the quantitative impact magnitude $c$.
The detailed taxonomy of event types, the corresponding annotation criteria,
and the dataset format are provided in Appendix~\ref{app:event_taxonomy}.

\subsection{Decision-Oriented Finetuning}
\textbf{Why}: 
(1) Purely supervised finetuning cannot ensure that model reasoning translates into market-consistent decisions.
(2) Existing reinforcement learning–based methods often rely on heuristic, additively designed rewards, where competing objectives may offset each other, limiting their ability to model economically meaningful trade‑offs.
\textbf{How}: Building on the event-centric dataset, we perform decision-oriented fine-tuning
to align model reasoning with executable trading actions. This stage follows a multi-step training paradigm.
We first apply supervised fine-tuning (SFT) to stabilize structured event reasoning,
and then perform reinforcement fine-tuning using Group Relative Policy Optimization (GRPO)
to directly optimize trading decisions. To ensure that reinforcement learning is guided by economically meaningful signals,
we design a structured reward mechanism, detailed below.

\subsubsection{Hierarchical-Gated Reward Modeling}
To bridge semantic event understanding with executable trading actions, we propose a
\emph{Hierarchical-Gated Reward Model} (HGRM) that provides structured supervision for reinforcement-based fine-tuning.
For a given stock $i$ at event time $t_0$, the model generates a response that is interpreted as a composite prediction
$\hat{y} = \{\hat{c}, \hat{d}, \hat{s}, \hat{e}\}$, where
$\hat{c}$ denotes the predicted cumulative abnormal return (CAR),
$\hat{d}$ represents the predicted direction (positive, negative, neutral),
$\hat{s}$ indicates whether a trade should be executed (strong or weak).
and $\hat{e}$ corresponds to the inferred event type.


For model outputs, we first extract the predicted direction $\hat{d}$ and trading strength $\hat{s}$ from the generated response $\hat{y}$ when these attributes are explicitly stated.
If either $\hat{d}$ or $\hat{s}$ is absent, the missing component is independently inferred from the predicted CAR $\hat{c}$,
with $\hat{d} = \mathrm{sign}(\hat{c})$ and $\hat{s}$ determined using the same threshold $\tau$.

\subsubsection{Hard gate: Direction correctness.}
We first enforce a hard direction gate $g_{\mathrm{dir}} \in \{0,1\}$ to prevent spurious profits under incorrect market polarity.
The direction score is defined as
\begin{equation}
s_{\mathrm{dir}}(\hat{d}, d)=
\begin{cases}
1, & \hat{d} = d,\\
-\lambda_{\mathrm{dir}}, & \hat{d} = -d,\\
0, & \text{otherwise},
\end{cases}
\end{equation}

where $\lambda_{\mathrm{dir}} > 1$ assigns a stronger penalty to opposite directions than to ambiguous cases involving \texttt{neutral}.
When $s_{\mathrm{dir}} < 0$, we set $g_{\mathrm{dir}}=0$ to block all subsequent reward contributions and avoid rewarding trades with incorrect directional alignment.

\subsubsection{Soft gate: Event-type consistency.}
We introduce an event-type soft gate that discounts rewards when the predicted event category is incorrect:
\begin{equation}
s_{\mathrm{evt}}(\hat{e}, e)=
\begin{cases}
1, & \hat{e} = e,\\
-\lambda_{\mathrm{evt}}, & \hat{e} \neq e,\\
-\lambda_{\mathrm{miss}}, & \hat{e} = \emptyset,
\end{cases}
\qquad
m_{\mathrm{evt}}=
\begin{cases}
1, & \hat{e} = e \ \\
\alpha, & \text{otherwise},
\end{cases}
\label{eq:event_gate}
\end{equation}

Here, $s_{\mathrm{evt}}(\hat e, e)$ provides a signed supervision signal for event-type correctness,
while $m_{\mathrm{evt}}$ acts as a multiplicative discount on trading rewards.
The penalty parameters $\lambda_{\mathrm{evt}} > 0$ and $\lambda_{\mathrm{miss}} > 0$ control the
strength of penalization for incorrect and missing event-type predictions (i.e., $\hat e=\emptyset$), respectively.
The discount factor $\alpha \in (0,1)$ reduces the contribution of profit-based rewards when the predicted event type
does not match the ground truth, encouraging consistency between the understanding of the event and the trading outcomes.

\subsubsection{Trading reward: cost-aware PnL with strength regularization.}
To reflect economically meaningful outcomes, we define a cost-aware single-event trading payoff based
on the realized CAR $c$:
\begin{equation}
\mathrm{PnL}(\hat{d},c)=
\begin{cases}
c-\kappa, & \hat{d}=\texttt{positive},\\
-c-\kappa, & \hat{d}=\texttt{negative},\\
0, & \hat{d}=\texttt{neutral},
\end{cases}
\label{eq:pnl}
\end{equation}
where $\kappa$ denotes the transaction cost.
The payoff is activated only when the predicted trading strength is $\hat{s}=\texttt{strong}$, scaled by the event-type discount factor $m_{\mathrm{evt}}$ and the realized profit and loss $\mathrm{PnL}(\hat{d}, c)$, and set to zero otherwise.
To stabilize reinforcement learning, the resulting reward is clipped to a symmetric range $[-\rho, \rho]$, where $\rho>0$ is a predefined bound. We denote the resulting clipped profit-based reward as $r_{\mathrm{pnl}} = \mathrm{clip}(m_{\mathrm{evt}}\cdot \mathrm{PnL}(\hat{d},c),-\rho,\rho)$.


In addition, we regularize the predicted trading strength to discourage degenerate strategies.
False-positive decisions (predicting \texttt{strong} when no trade is warranted) and false-negative
decisions (predicting \texttt{weak} when profitable opportunities exist) are penalized with asymmetric
costs, preventing the model from collapsing to always-trade or never-trade behaviors.  

\subsubsection{Magnitude shaping and process reward.}
To provide fine-grained supervision beyond direction,
we add a magnitude shaping term when the direction is not blocked:
\begin{equation}
r_{\mathrm{mag}}=
\exp\!\left(-\frac{|\hat{c}-c|}{\sigma}\right),
\label{eq:mag}
\end{equation}
where $\sigma$ controls the tolerance scale.
We also include a lightweight process reward $r_{\mathrm{proc}}$ that evaluates the presence of required reasoning sections and penalizes overly long responses and excessive self-questioning.

\subsubsection{Overall reward.}
We compose the final HGRM reward hierarchically, where higher-level gates control whether lower-level rewards are activated:
\begin{equation}
\begin{aligned}
R
&= w_{\mathrm{dir}}\, s_{\mathrm{dir}} \\
&\quad + g_{\mathrm{dir}}\Big(
    w_{\mathrm{evt}}\, s_{\mathrm{evt}}
    + w_{\mathrm{pnl}}\, r_{\mathrm{pnl}}
    + w_{\mathrm{mag}}\, r_{\mathrm{mag}}
    + w_{\mathrm{proc}}\, r_{\mathrm{proc}}
\Big).
\end{aligned}
\end{equation}
where $g_{\mathrm{dir}} \in \{0,1\}$ blocks lower-level rewards when the predicted direction is opposite to the realized movement (i.e., $s_{\mathrm{dir}}<0$).
The Algorithm~\ref{alg:hgrm} formalizes the hierarchical gated reward modeling procedure. The corresponding training configurations and implementation details are provided in Appendix~\ref{app:training_details}.

\section{Experiment Results}

\subsection{Experiment Setup}
Given a universe of stocks $\mathcal{S}$, for any stock $s \in \mathcal{S}$ on a trading day $t$,
we evaluate a long--short trading strategy driven by Janus-Q.
The trading process is defined as follows.
\textbf{(1) Signal Generation.}
Janus-Q analyzes all news events released between the market open at $9{:}30$ on day $t$
and the market open at $9{:}30$ on day $t{+}1$, and produces a directional signal
$\gamma_{s,t} \in \{\text{Long}, \text{Short}, \text{Hold}\}$ for each stock $s$.
\textbf{(2) Event-weighted Signal Aggregation.}
On each trading day $t$, news signals are aggregated at the portfolio level using event-type weights $w_k$
estimated from historical post-event abnormal returns.
The daily budget is allocated across event types and evenly split within each type using only information available before day $t{+}1$ market open.
\textbf{(3) Entry Rule.}
If the aggregated signal $\gamma_{s,t}$ indicates a long (short) position,
we initiate a long (short) trade at the opening price $o_{s,t+1}$ on day $t{+}1$.
\textbf{(4) Exit Rule.}
The position is held until the last trading day within the subsequent two trading days, denoted as $\tau(t)$, at which point the stock is sold at the closing price $c_{s,\tau(t)}$.


\subsection{Evaluation Objective}

We conduct a comprehensive evaluation to assess the effectiveness of Janus-Q in event-driven trading.
Our evaluation is designed to examine both overall trading performance and the contribution of key design components,
with a particular focus on decision alignment, reward modeling, and human-consistent understanding.
To this end, we structure our experiments around the following research questions (RQs):

\noindent$\bullet$~\textbf{RQ1:} Does Janus-Q consistently outperform market indices and competitive model baselines
in terms of trading performance and decision accuracy?

\noindent$\bullet$~\textbf{RQ2:} How does each core component of Janus-Q contribute to its overall trading performance?

\noindent$\bullet$~\textbf{RQ3:} How effective are diversified reward objectives in improving learning stability
and decision-oriented trading performance?

\noindent$\bullet$~\textbf{RQ4:} To what extent does Janus-Q align with human judgments 
in interpreting financial events?

\subsection{Dataset \& Metrics}
We evaluate our model on a multi-source dataset that integrates both textual and financial information.
Raw news articles are collected from the Datayes platform\footnote{\url{https://www.datayes.com}},
spanning the period from January 1, 2023 to January 25, 2025.
Corresponding stock price data for backtesting are retrieved from Tushare\footnote{\url{https://tushare.pro}},
covering the period from January 1, 2023 to February 6, 2025.
All data are chronologically split into a 4:4:1:1 ratio for historical statistics, training, validation, and testing.
Additionally, we incorporate corporate profiles, including industry category and market share—sourced from the Wind platform\footnote{\url{https://www.wind.com.cn}} to support semantic enrichment.

To evaluate the performance of our model, we adopt six metrics: 
\textbf{Mean Absolute Error (MAE)}, \textbf{Root Mean Square Error (RMSE)},
\textbf{Direction Accuracy (DA)}, 
\textbf{Event Type Accuracy (ETA)}, 
\textbf{Sharpe Ratio (SR)},  and \textbf{Maximum Drawdown (MDD)}.
Together, these metrics provide a comprehensive view of both predictive quality and practical investment performance of our model.

\subsection{Baseline}
Our proposed approach is evaluated against four categories of baselines:

\noindent$\bullet$~\textbf{Market Indices.} 
Market indices serve as standard passive investment benchmarks. We report results on several representative indices, including the \textbf{CSI 300}, \textbf{CSI 500}, and \textbf{CSI 1000}, which collectively cover large-, mid-, and small-cap segments of the Chinese equity market.

\noindent$\bullet$~\textbf{Time-Series-Oriented LLM.} 
We include a set of LLM specifically designed for modeling temporal and sequential data, which adapt LLM architectures to time-series forecasting tasks. Representative models in this category include Time-MQA \cite{kong2025time}, ChatTS-14B \cite{xie2024chatts}, and TimeMaster \cite{zhang2025timemaster}. 

\noindent$\bullet$~\textbf{Financial Domain-Specific LLM.} 
We evaluate several open-source language models pre-trained or fine-tuned on financial corpora, including FinMA~\cite{xie2023pixiu}, DISC-FinLLM~\cite{chen2023disc}, and Stock-Chain~\cite{li2024alphafin}. 
These models are designed to capture financial semantics and domain-specific patterns, enabling direct application to market analysis and trading-related evaluation.

\noindent$\bullet$~\textbf{General-Purpose LLM.} 
We further include widely used general-purpose language models, including QwQ-32B \cite{yang2025qwen3}, Claude-3-Haiku \cite{rahman2024automatic}, GPT-4o-mini~\cite{hurst2024gpt}, DeepSeek-v3.1-nex-n1 \cite{cai2025nex}, Grok-3-mini-beta \cite{jiang2025makes}, Qwen2.5-7B \cite{team2024qwen2}, and Gemini-2.5-flash \cite{comanici2025gemini}, which are not specifically tuned for financial tasks.

\begin{figure*}[ht] 
    \centering
    \includegraphics[width=0.8\textwidth, height=0.3\textheight,
    trim=40pt 10pt 40pt 10pt, 
    clip]{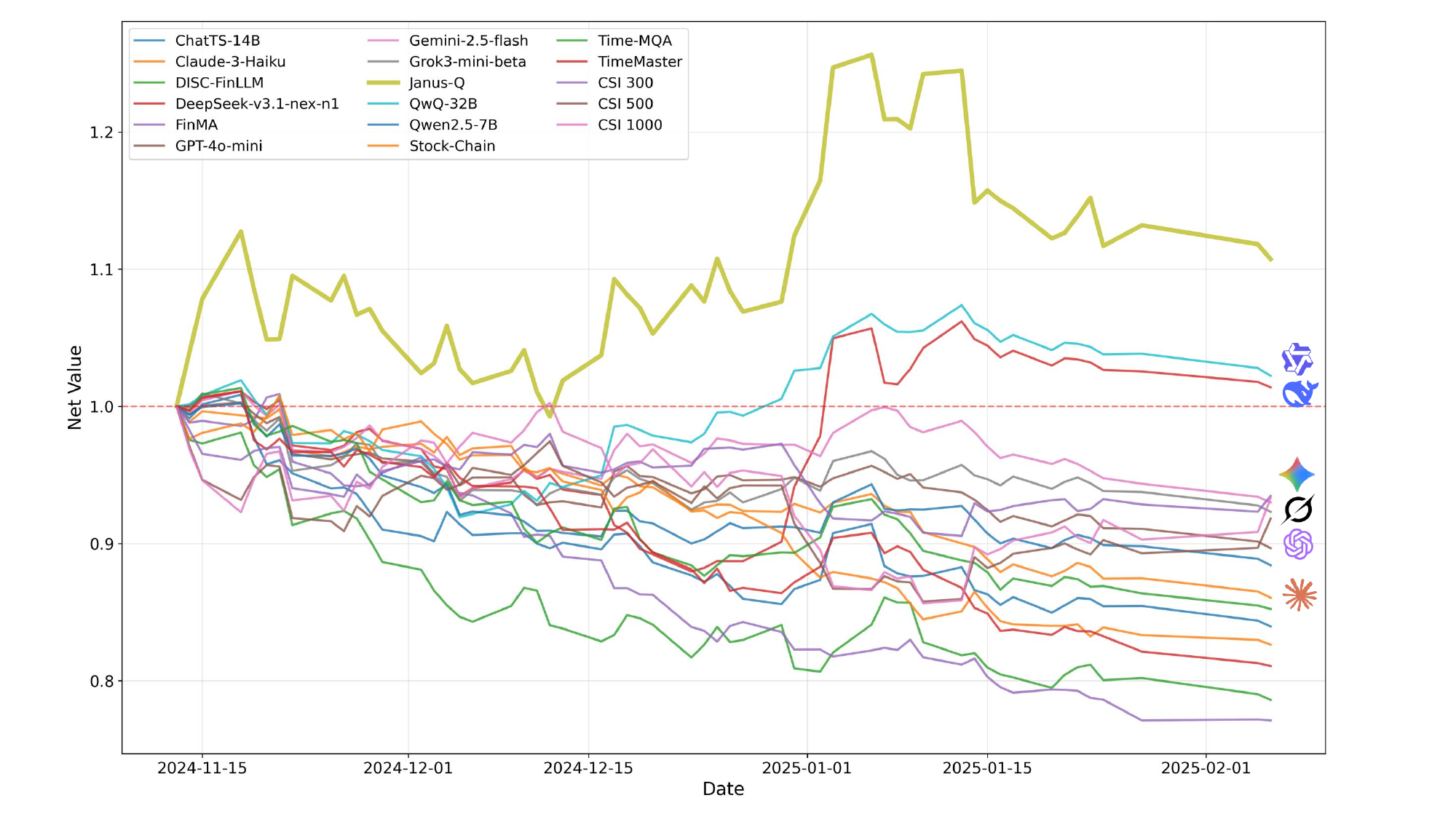} 
    \caption{Cumulative returns of each baseline strategy on the financial news dataset from November 12, 2024 to February 6, 2025. The figure shows the net asset value (NAV) curves over the backtesting period.}
    \label{fig:nav_curve} 
\end{figure*}

\begin{table*}[t]
\centering
\caption{Performance comparison across Market Indices, Vanilla LLM, Time-aware LLM, and Financial LLM.}
\label{tab:model_comparison}
\renewcommand{\arraystretch}{1.00}
\setlength{\tabcolsep}{12pt}
\begin{tabular}{l|l|cccccc}
\toprule
& \textbf{Model} & \textbf{MAE↓} & \textbf{RMSE↓} & \textbf{DA↑} & \textbf{ETA↑} & \textbf{SR↑}  & \textbf{MDD↓}\\
\midrule\midrule
\multirow{3}{*}{Market Indices}
 & CSI 300   & -- & -- & -- & -- & -1.8027 & 0.0945 \\
 & CSI 500 & -- & -- & -- & -- &  -1.6074 & 0.1423\\
 & CSI 1000 & -- & -- & -- & -- & -0.1036 & 0.1456 \\
\midrule\midrule
\multirow{3}{*}{TimeLLM}
 & Time-MQA   & 0.0427 & 0.2651 & 0.4732 & 0.5958 & -4.2184 & 0.1589 \\
 & ChatTS-14B & 0.0932 & 0.3271 & 0.4995 & 0.6223 & -4.2827 & 0.1675 \\
 & TimeMaster & 0.0472 & 0.1015 & 0.4472 & 0.6263 & -5.9237 & 0.1981 \\
\midrule\midrule
\multirow{3}{*}{FinLLM}
 & Stock-Chain  & 0.1352 & 0.3226 & 0.4510 & 0.4364 & -5.3581 & 0.1800 \\
 & FinMA        & 0.0947 & 0.2531 & 0.4550 & 0.5714 & -6.9668 & 0.2358 \\
 & DISC-FinLLM  & 0.0707 & 0.1672 & 0.4286 & 0.5608 & -4.6307 & 0.2139 \\
\midrule\midrule
\multirow{7}{*}{Vanilla LLM}
 & QwQ-32B              & 0.0437 & 0.0839 & 0.4674 & 0.7080 & 0.6481 & 0.0979 \\
 & Claude-3-Haiku       & 0.0384 & 0.0557 & 0.4547 & 0.4884 & -5.4698 & 0.1395 \\
 & GPT-4o-mini          & 0.0387 & 0.0558 & 0.4700 & 0.7508 & -3.8759 & 0.1060 \\
 & DeepSeek-v3.1-nex-n1 & 0.0541 & 0.1214 & 0.4365 & 0.6811 & 0.3710 & 0.1300 \\
 & Grok-3-mini-beta     & 0.0369 & 0.0559 & 0.4795 & 0.5417 & -2.2698 & 0.0855 \\
 & Qwen2.5-7B           & 0.0377 & 0.0551 & 0.4114 & 0.4879 & -3.9154 & 0.1177 \\
 & Gemini-2.5-flash     & 0.0534 & 0.0730 & 0.4276 & 0.6797 & -2.0899 & \textbf{0.0805} \\
\midrule\midrule
\multirow{1}{*}{Finetuning LLM}
& \cellcolor{gray!15}\textbf{Janus-Q (Ours)}
& \cellcolor{gray!15}\textbf{0.0349}
& \cellcolor{gray!15}\textbf{0.0541}
& \cellcolor{gray!15}\textbf{0.5869}
& \cellcolor{gray!15}\textbf{0.8009}
& \cellcolor{gray!15}\textbf{1.3088}  
& \cellcolor{gray!15}0.1196
\\

\bottomrule
\end{tabular}
\end{table*}

\subsection{Experimental Results}

\subsubsection{Baseline Comparison (RQ1)}
We compare Janus-Q against 16 baseline methods across six general evaluation metrics. Figure~\ref{fig:nav_curve} depicts a market regime characterized by a correction phase, followed by a broad pullback and consolidation, reflecting the waning influence of favorable Chinese fiscal policy signals and subsequent index-level corrections.
During this period, all three benchmark indices fail to sustain positive risk-adjusted returns, with Sharpe Ratios remaining negative throughout. Most time-aware models and financial LLM exhibit similar patterns, either enduring persistent drawdowns or demonstrating weak, oscillatory NAV trajectories that closely mirror the underlying market dynamics. Among general-purpose LLM, only QwQ-32B and DeepSeek-v3.1-nex-n1 achieve positive Sharpe Ratios; however, their NAV curves remain highly volatile and lack sustained upward momentum.
In contrast, Janus-Q exhibits a distinctly different trajectory. It effectively captures the sharp upswing observed in late December and maintains a stable and consistent growth trend thereafter. By the end of the evaluation window, Janus-Q achieves the highest cumulative return among all compared methods, highlighting its superior ability to adapt to shifting market conditions and translate predictive precision into tangible trading gains.

The quantitative results in Table~\ref{tab:model_comparison} corroborate these findings. 
Janus-Q consistently outperforms all baselines on both decision-level and trading-level metrics. 
It achieves the lowest CAR prediction error, reducing MAE by 18.3\% relative to the best time-aware model and by 50.6\% over the strongest financial LLM baseline; even against the best vanilla LLM, Janus-Q still lowers error by 5.4\%. 
Its direction accuracy further exceeds the best time-aware, financial, and vanilla LLM baselines by 17.5\%, 29.0\%, and 22.4\%, respectively.
Notably, domain-specific FinLLM underperform most vanilla LLM, likely due to task–domain mismatch during finetuning and architectural or context-length constraints that hinder effective modeling of long, complex news narratives.
Meanwhile, time-aware models remain inferior in backtesting performance despite explicitly incorporating textual features. 
This indicates that merely injecting textual cues as auxiliary inputs is insufficient for robust trading, reinforcing the value of directly optimizing the event-to-decision mapping. 
These decision-level gains translate into significantly better trading outcomes. 
Janus-Q attains a Sharpe Ratio of 1.3088, more than doubling that of the runner-up QwQ-32B and yielding a relative improvement above 102.0\%. 
In contrast, both time-aware and financial LLM baselines exhibit negative Sharpe Ratios over the same test horizon. 
Although some baselines achieve marginally lower drawdowns, this comes at the cost of substantially reduced returns, whereas Janus-Q attains a more favorable balance between profitability and stability with superior risk-adjusted performance.

\subsection{Ablation Study}

\subsubsection{Effectiveness of Each Component (RQ2)}

Table~\ref{tab:ablation module component} reports the ablation study evaluating the contribution of each main component in Janus-Q. Removing any component consistently degrades performance, confirming that the framework’s effectiveness arises from the synergy among its core modules. Among all variants, eliminating supervised fine-tuning leads to the most pronounced deterioration, with direction accuracy declining by over 14\% and trading performance shifting from a robustly positive regime to a distinctly negative Sharpe Ratio. This finding underscores its fundamental role in establishing a reliable decision-making foundation. In comparison, removing reinforcement optimization causes a moderate yet consistent decline of approximately 13\% in Sharpe Ratio, suggesting that reinforcement learning primarily fine-tunes and enhances an already acquired policy rather than replacing supervised learning. Ablating CAR supervision or company-level information results in smaller but still noticeable reductions in both predictive accuracy and profitability, demonstrating that market impact cues and firm-specific context make meaningful contributions to effective event-driven trading.

\begin{table}[t]
\centering
\caption{Ablation study of Janus-Q. Each variant removes a single source-level or structural component from the full model.}
\label{tab:ablation module component}
\renewcommand{\arraystretch}{1.00}
\setlength{\tabcolsep}{8pt}
\begin{tabular}{l|cccc}
\toprule
\textbf{Methods} & \textbf{MAE} & \textbf{DA} & \textbf{ETA} & \textbf{SR}\\
\midrule
w/o CAR    & 0.0387 & 0.5261 & 0.7796 & 0.8690 \\
w/o Co. Info  & 0.0353 & 0.5464 & 0.7790 & 0.9608 \\
w/o SFT   & 0.0381 & 0.4429 & 0.6771 & -5.2848 \\
w/o GRPO  & 0.0355 & 0.5459 & 0.7881 & 1.1330\\
\midrule
\textbf{Janus-Q (Full)} & \textbf{0.0349} & \textbf{0.5869} & \textbf{0.8009} & \textbf{1.3088}\\
\bottomrule
\end{tabular}
\end{table}

\subsubsection{Effectiveness of Diversified Reward Objectives (RQ3)}

\begin{table}[t]
\centering
\caption{Ablation study of diversified reward objectives in HGRM.
Each variant removes a single reward component from the full model.}
\label{tab:ablation reward model}
\renewcommand{\arraystretch}{1.00}
\setlength{\tabcolsep}{8pt}
\begin{tabular}{l|cccc}
\toprule
\textbf{Methods} & \textbf{MAE} & \textbf{DA} & \textbf{ETA} & \textbf{SR}\\
\midrule
w/o Event Type    & \textbf{0.0345} & 0.5819 & 0.7817 & 1.2107\\
w/o Direction    & 0.0357 & 0.5589 & 0.7896 & 1.2788 \\
w/o Magnitude  & 0.0347 & 0.5653 & 0.7843 & 1.1558 \\
w/o PNL   & 0.0354 & 0.5799 & 0.7868 & 1.1953 \\

\midrule
\textbf{HGRM (Full)} & 0.0349 & \textbf{0.5869} &  \textbf{0.8009} & \textbf{1.3088} \\
\bottomrule
\end{tabular}
\end{table}

Table~\ref{tab:ablation reward model} examines the impact of removing individual reward objectives from the full HGRM.
Overall, all ablated variants underperform the full model, indicating that diversified reward objectives play complementary roles in learning stable and effective trading policies.
Removing the direction-related objective leads to the most pronounced drop in decision quality,
with direction accuracy decreasing by over 4.8\%,
while the impact on trading performance remains comparatively moderate, likely because the model still captures a large fraction of high-impact signals.
In contrast, removing event-type supervision slightly reduces prediction error
and leads to a modest decline in trading performance,
as the absence of event-type signals affects position weighting
and weakens the allocation of capital across heterogeneous events.
A detailed analysis of this effect is provided in Appendix,
as illustrated in Figure~\ref{fig:event_weights_barplot} and Figure~\ref{fig:two type weight}.
Eliminating magnitude or profit-and-loss objectives results in more pronounced performance degradation,
reducing the Sharpe Ratio by approximately 11.7\% and 8.7\%, respectively.
In comparison, the full HGRM achieves the highest decision accuracy and the strongest trading performance,
demonstrating that jointly optimizing diversified reward objectives is critical for stable learning
and decision-oriented event-driven trading.

\subsection{Case Study}

\subsubsection{Human Alignment in Event Interpretation (RQ4)}

\begin{figure}[t]
    \centering
    \vspace{-6px}
    \begin{subfigure}[t]{0.48\textwidth}
        \centering
        \includegraphics[width=\textwidth]{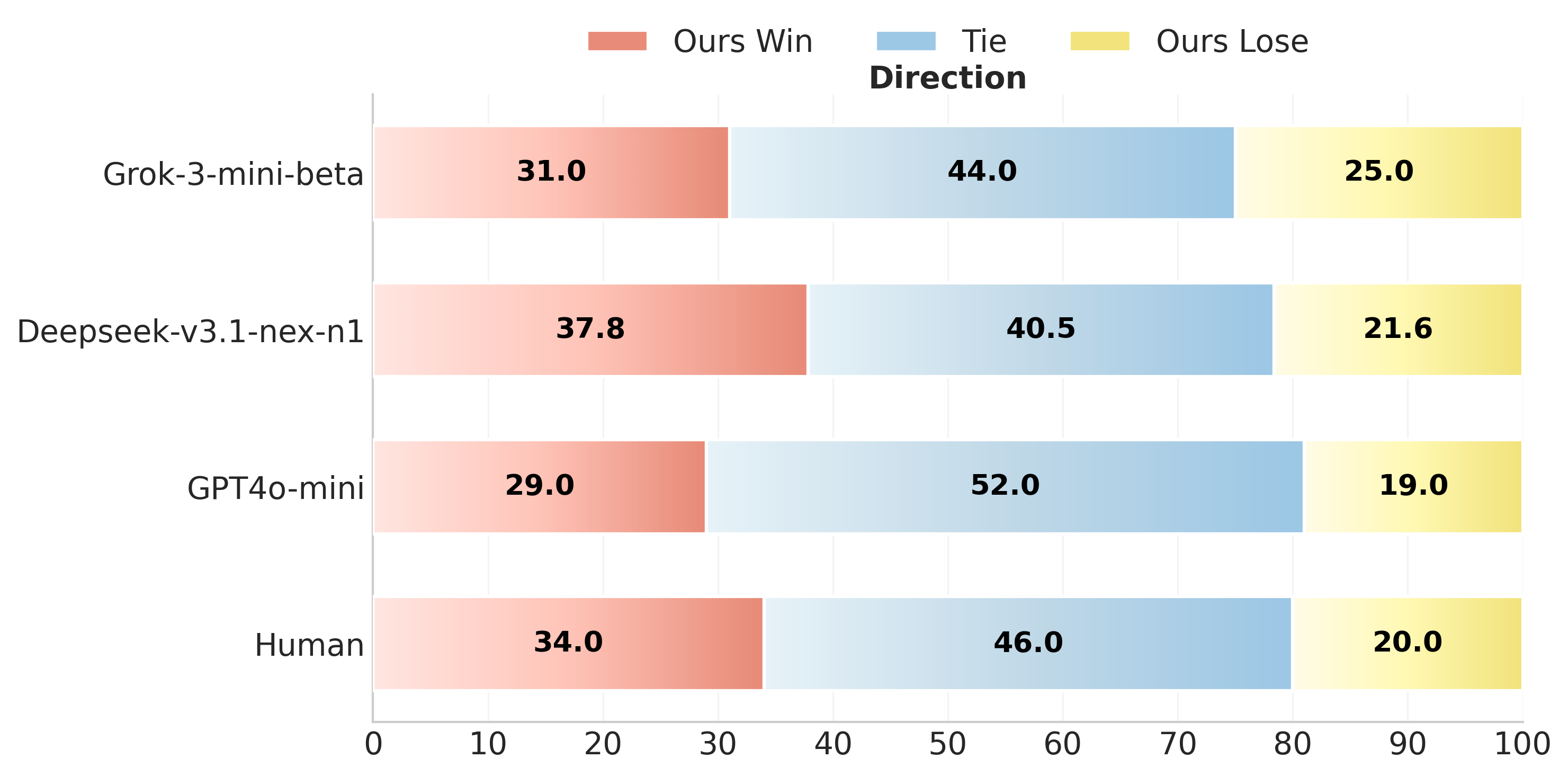}
    \end{subfigure}
    \hfill
    \begin{subfigure}[t]{0.48\textwidth}
        \centering
        \includegraphics[width=\textwidth]{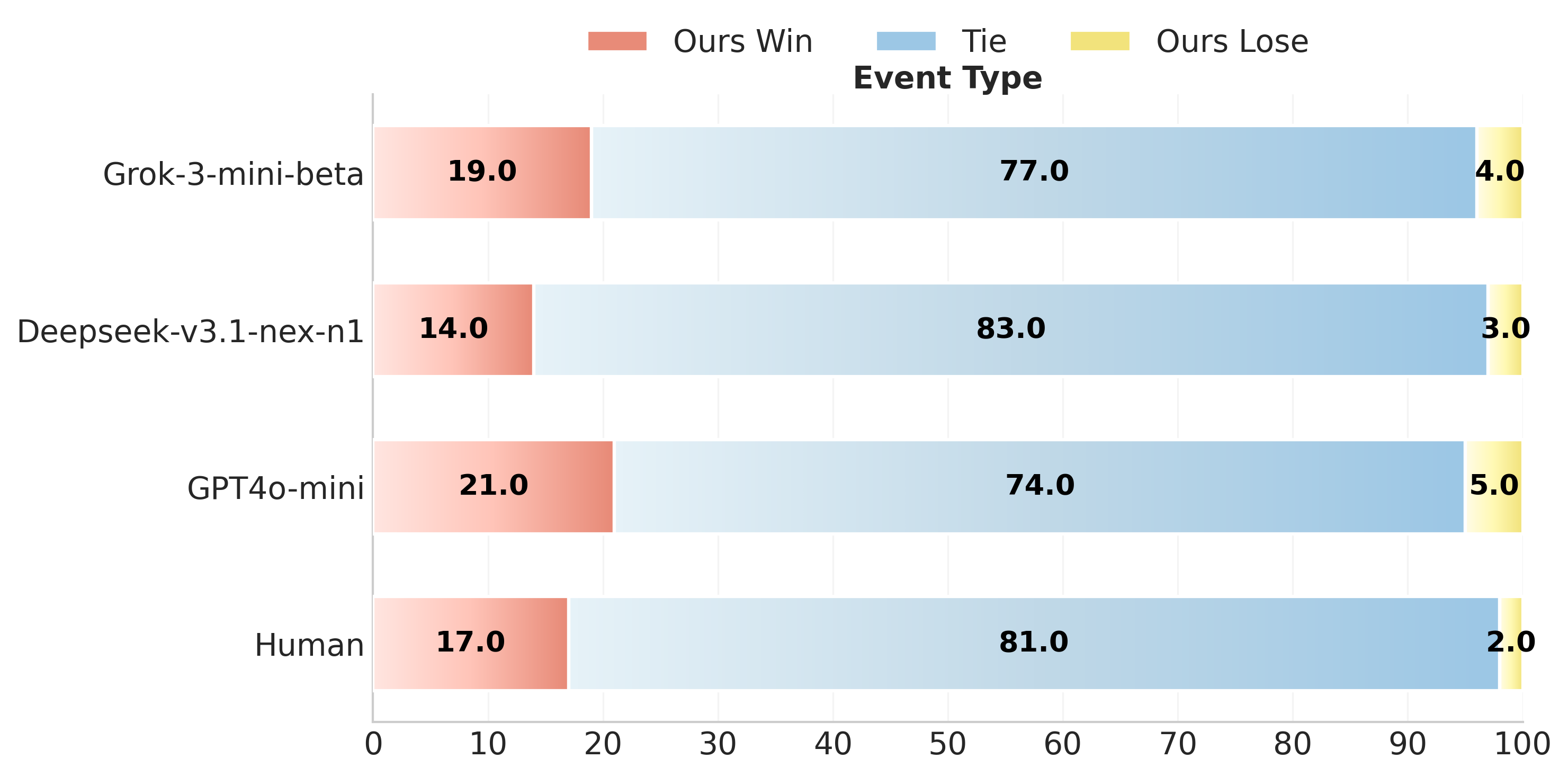}
    \end{subfigure}

    \vspace{-5px}
    \caption{Visualization of comparative evaluation results between humans and models.}
    \vspace{-5px}
    \label{fig:human_model_eval}
\end{figure}

Figure~\ref{fig:human_model_eval} reports a direct comparison among Janus-Q, strong LLM baselines, and human evaluations in event interpretation. The evaluation is conducted on 200 randomly selected samples from the test dataset. 
To obtain a representative measure of human judgment, we recruit three participants with varying levels of financial expertise, including a finance major student, a securities analyst, and a CFA certificant. This design captures a spectrum of professional backgrounds and mitigates potential bias from individual knowledge gaps. Their judgments are aggregated through majority voting to form a balanced human consensus benchmark.
For direction prediction, Janus-Q demonstrates substantial agreement with both humans and other models, achieving tie rates between 40.5\% and 52.0\%, win rates of up to 37.8\% against DeepSeek-v3.1-nex-n1, and loss rates not exceeding 25.0\%. The alignment becomes even more pronounced in event-type understanding, where tie cases dominate across comparisons, accounting for 74.0\% to 83.0\%, and loss rates remain below 5.0\% for all counterparts, including human judges. In addition, Janus-Q secures non-trivial win rates ranging from 14.0\% to 21.0\%, suggesting that its deviations from human judgment are infrequent and occur in a controlled manner rather than reflecting systematic semantic discrepancies. 
Overall, these results show that Janus-Q demonstrates a higher level of accuracy in event interpretation compared to the average performance of human evaluators, as variations in human judgment may arise from the subjective nature of the evaluation. Janus-Q also retains the flexibility required for decision-oriented and context-sensitive trading behavior.

\section{Conclusion}

This paper investigates event‑driven trading as an alternative to conventional time‑series‑centric formulations and demonstrates that explicitly modeling financial news events as primary decision units leads to more reliable and interpretable trading behavior.
We propose Janus‑Q, a two‑stage event‑driven trading framework that directly maps financial events to trading decisions.
The framework first constructs a large‑scale event‑centric dataset that serves as a unified benchmark for event‑level market impact analysis by linking fine‑grained event semantics with empirically grounded market reactions.
Building on this foundation, Janus‑Q adopts a multi‑step training paradigm that combines supervised reasoning alignment with reinforcement fine‑tuning under a hierarchical gated reward scheme.
Extensive experiments show that Janus‑Q consistently surpasses both market indices and strong large‑language‑model baselines in decision accuracy and trading performance, highlighting the importance of aligning language model reasoning with financially meaningful objectives.
In future work, we plan to extend this framework to support finer‑grained event structures, richer multimodal inputs, and cross-market backtesting, further advancing event‑driven learning for real‑world financial trading.



\newpage

\bibliographystyle{ACM-Reference-Format}
\bibliography{main}

\appendix

\newpage

\section{Appendix}

\begin{table}[t]
\centering
\caption{SFT training hyperparameters.}
\label{tab:sft_hparams}
\setlength{\tabcolsep}{6pt}
\renewcommand{\arraystretch}{1.08}
\begin{tabularx}{\columnwidth}{@{} l X @{}}
\toprule
\textbf{Hyperparameter} & \textbf{Value / Setting} \\
\midrule

\multicolumn{2}{@{}l}{\textbf{Data Configuration}} \\
\addlinespace[2pt]
Max sequence length & 1400 \\
Dataset shuffling & True \\
\addlinespace[4pt]
\midrule

\multicolumn{2}{@{}l}{\textbf{Training Configuration}} \\
\addlinespace[2pt]
Epochs & $\{1, 2, 3\}$ \\
Train micro-batch size per device & 8 \\
Gradient accumulation steps & 4 \\
Total effective batch size & $8 \times 4 \times N_{\mathrm{GPU}}$ \\
Precision & BF16 \\
Gradient checkpointing & True\\
Evaluation steps & 100 \\
Logging steps & 10 \\
\addlinespace[4pt]
\midrule

\multicolumn{2}{@{}l}{\textbf{Optimizer \& LR Scheduler}} \\
\addlinespace[2pt]
Optimizer & torch.optim.AdamW \\
Learning rate & $5\times10^{-6}$ \\
Warmup ratio & 0.2 \\
Weight decay & 0.01 \\
AdamW betas & 0.9, 0.999  \\
\addlinespace[4pt]
\midrule

\multicolumn{2}{@{}l}{\textbf{LoRA Configuration}} \\
\addlinespace[2pt]
LoRA rank $r$ & 8 \\
LoRA alpha $\alpha$ & 16 \\
LoRA dropout & 0.1 \\
Task type & \texttt{CAUSAL\_LM} \\
\addlinespace[4pt]
\bottomrule
\end{tabularx}
\end{table}

\begin{table}[t]
\centering
\caption{GRPO training hyperparameters.}
\label{tab:grpo_hparams}
\setlength{\tabcolsep}{6pt}
\renewcommand{\arraystretch}{1.08}
\begin{tabularx}{\columnwidth}{@{} l X @{}}
\toprule
\textbf{Hyperparameter} & \textbf{Value / Range} \\
\midrule

\multicolumn{2}{@{}l}{\textbf{Training Configuration}} \\
\addlinespace[2pt]
Epochs & $\{1, 2, 3\}$ \\
Batch size & $\{1, 2, 4\}$ \\
Learning rate & $\{1\mathrm{e}{-6}, 2\mathrm{e}{-6}, 5\mathrm{e}{-6}\}$ \\
Max grad norm & $\{0.5, 1.0\}$ \\
\addlinespace[4pt]
\midrule

\multicolumn{2}{@{}l}{\textbf{Ref Configuration}} \\
\addlinespace[2pt]
PPO epochs & $\{1, 2\}$ \\
Samples per prompt & $\{2, 4, 8\}$ \\
Initial KL coefficient & $\{0.05, 0.1, 0.2\}$ \\
Target KL & $\{0.01, 0.02\}$ \\
Clip range & $\{0.1, 0.2\}$ \\
Normalize advantage by std & \{True, False\} \\
\addlinespace[4pt]
\midrule

\multicolumn{2}{@{}l}{\textbf{Data Configuration}} \\
\addlinespace[2pt]
Max prompt length &  $\{3000, 4196\}$ \\
Max response length & $\{4196, 8192\}$ \\
Data Shuffling & \{True, False\} \\
\addlinespace[4pt]
\midrule

\multicolumn{2}{@{}l}{\textbf{System Configuration}} \\
\addlinespace[2pt]
Number of GPUs & $\{1, 2, 4, 8\}$ \\
Number of nodes & $\{1, 2\}$ \\
vLLM inference & \{True, False\} \\
DeepSpeed ZeRO stage & \{None, ZeRO-2, ZeRO-3\} \\
Mixed precision & \{FP16, BF16\} \\
\bottomrule
\end{tabularx}
\end{table}

\subsection{Hyperparameters \& Datasets \& Metrics}

\subsubsection{Training details.}
\label{app:training_details}

We adopt a two-phase training strategy consisting of supervised fine-tuning (SFT) followed by
reinforcement fine-tuning (RFT) to stabilize event reasoning and optimize trading decisions.
All experiments are conducted on $8\times$ NVIDIA A100 GPUs (40GB).

\noindent $\bullet$ \textbf{Supervised fine-tuning.}
We first apply SFT with LoRA adaptation to initialize a reasoning-aware policy that predicts
event semantics and market impact in a stable manner.
The detailed hyperparameter configuration for SFT is summarized in Table~\ref{tab:sft_hparams}.

\noindent $\bullet$ \textbf{Reinforcement fine-tuning.}
Building on the initialized SFT policy, we further perform reinforcement fine-tuning using
Group Relative Policy Optimization (GRPO) together with the proposed hierarchical-gated reward model,
to directly optimize decision-oriented trading objectives.
The detailed hyperparameter settings for GRPO are reported in Table~\ref{tab:grpo_hparams}.

\begin{table}[t]
\centering
\caption{Summary of event-centric dataset.}
\label{tab:dataset_summary}
\setlength{\tabcolsep}{6pt}
\renewcommand{\arraystretch}{1.08}
\begin{tabularx}{\columnwidth}{@{} l X @{}}
\toprule
\textbf{Field} & \textbf{Description} \\
\midrule
News & Raw financial news article  \\
Event time ($t_0$) & Publication timestamp of the news \\
StockInfo ($i$) & Associated traded asset and its basic profile \\
Event type ($e$) & Sematic category of event \\
Direction ($d$) & Market reaction direction: positive / negative / neutral \\
Strength ($s$) & Trading strength indicator: strong / weak \\
CAR ($c$) & Cumulative abnormal return \\
\bottomrule
\end{tabularx}
\end{table}

\subsubsection{Dataset}
\label{app:event_taxonomy}
We construct a large-scale Chinese equity dataset covering 5,282 A-share stocks, integrating structured market data with multi-source textual information. Formally, the dataset consists of a collection of event instances,
each corresponding to a financial news event associated with a specific
traded asset and timestamp.
each event instance is represented as a structured record:
$\mathcal{D}_{\text{event}} = \{ (\text{News}, t_0, \text{StockInfo}, e, d, s, c) \}$,
following the format defined in Table~\ref{tab:dataset_summary}.
Here, \textit{News} denotes the original news text,
$t_0$ the event timestamp,
\textit{StockInfo} the associated traded asset,
$e$ the fine-grained event type,
$d$ the directional label,
$s$ the strength indicator,
and $c$ the event-driven cumulative abnormal return.

Daily price and volume data are obtained from Tushare, while textual signals are collected from the Datayes platform, complemented by firm-level profile information from Wind.
To support stable event statistics and avoid information leakage, the dataset is chronologically partitioned into a historical window for statistical estimation, followed by non-overlapping training, validation, and test splits.
Detailed data ranges and split configurations are summarized in Table~\ref{tab:dataset_description}.

\begin{figure}[t]
    \centering
    \includegraphics[width=1.0\linewidth,
    trim=15pt 0pt 3pt 5pt,
    clip]{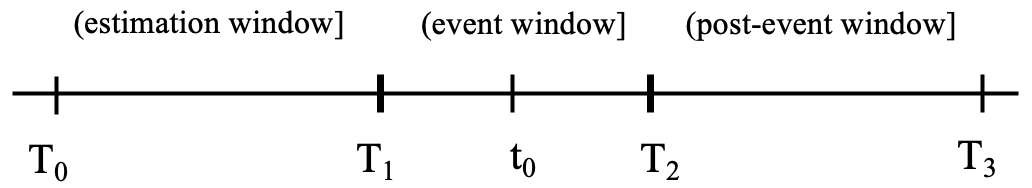}
    \caption{Temporal structure of the event study framework.}
    \label{fig:timeline}
\end{figure}

To characterize temporal dependencies around each event, we adopt a standard event‑study timeline consisting of three individual intervals: an estimation window $\mathcal{T}{\text{est}} = (T_0, T_1]$, an event window $\mathcal{T}{\text{evt}} = (T_1, T_2]$, and a post‑event window $\mathcal{T}{\text{post}} = (T_2, T_3]$, as illustrated in Figure~\ref{fig:timeline}. The estimation window precedes the event timestamp $t_0$ and provides a stable basis for estimating event‑free normal returns, ensuring that model calibration remains unaffected by upcoming market reactions. A short lag between $\mathcal{T}{\text{est}}$ and $\mathcal{T}{\text{evt}}$ is introduced to mitigate possible information leakage from early disclosures. The event window $\mathcal{T}{\text{evt}}$ captures immediate abnormal price responses, while the post‑event window $\mathcal{T}{\text{post}}$ extending to $T_3$ reflects subsequent return drift and adjustment dynamics. This study focuses on market reactions within $\mathcal{T}{\text{evt}}$ and does not model long‑horizon price dynamics within $\mathcal{T}{\text{post}}$, where returns generally revert toward their normal state.

Beyond scale, the dataset is distinguished by its rich semantic structure and high-quality financial annotations.
Each event is categorized into a diverse set of event types, including \emph{personal behavior}, \emph{equity change}, \emph{asset change}, \emph{dividend}, \emph{risk warning}, \emph{financing}, \emph{financial status}, \emph{violation}, \emph{industry}, and \emph{rating adjustment}.
This taxonomy is designed to cover a broad spectrum of firm-specific actions and external market signals, capturing heterogeneous information sources that are known to drive stock price movements.
Moreover, each event is associated with \underline{sentiment labels} and paired with its
\underline{cumulative abnormal return}, enabling direct supervision at both the semantic
and economic levels.
This dataset is intended to support future academic research on financial event understanding, market impact analysis, and event-driven modeling.

\begin{table*}[t]
\centering
\caption{Dataset Statistics}
\label{tab:dataset_description}
\setlength{\tabcolsep}{10pt}
\renewcommand{\arraystretch}{1.15}
\begin{tabular}{c c c c c c}
\toprule
\textbf{Country} & \textbf{Stock} & \textbf{Split} & \textbf{Date Range} & \textbf{Price} & \textbf{Text / Company Info} \\
\midrule
\multirow{4}{*}{CN} 
& \multirow{4}{*}{5282}
& Historical Stat & 2023/01/01 -- 2023/10/24 & \multirow{4}{*}{Tushare} & \multirow{4}{*}{Datayes / Wind} \\
& & Train & 2023/10/25 -- 2024/08/27 &  &  \\
& & Validation & 2024/08/28 -- 2024/11/11 &  &  \\
& & Test & 2024/11/12 -- 2025/01/25 &  &  \\
\bottomrule
\end{tabular}
\end{table*}

\subsubsection{Metrics}

To evaluate the performance of Janus-Q, we adopt a set of complementary metrics that jointly assess event-level prediction accuracy, decision correctness, and trading performance.

\textbf{Mean Absolute Error (MAE)}:
\[
\text{MAE} = \frac{1}{N} \sum_{i=1}^{N} \left| \hat{c}_i - c_i \right|
\]

\textbf{Root Mean Square Error (RMSE)}:
\[
\text{RMSE} = \sqrt{\frac{1}{N} \sum_{i=1}^{N} \left( \hat{c}_i - c_i \right)^2}
\]
where $\hat{c}_i$ is the predicted CAR, and $c_i$ is the true CAR.  MAE measures the average deviation between the predicted and true values of abnormal returns, capturing the overall accuracy of the predictions. RMSE assesses the sensitivity of the model to large errors, penalizing large deviations more heavily than MAE.

\textbf{Direction Accuracy (DA)}:
\[
\text{DA} = \frac{1}{N} \sum_{i=1}^{N} \mathbb{I} \left( \text{sign}(\hat{d}_i) = \text{sign}(d_i) \right)
\]
where $\mathbb{I}$ is the indicator function, $\hat{d}_i$ is the predicted direction, and $d_i$ is the true direction. DA evaluates whether the predicted direction of the market movement aligns with the actual result achieved, providing information on the accuracy of the model's direction prediction.

\textbf{Event Type Accuracy (ETA)}:
\[
\text{ETA} = \frac{1}{N} \sum_{i=1}^{N} \mathbb{I} \left( \hat{e}_i = e_i \right)
\]
where $\hat{e}_i$ is the predicted event type, and $e_i$ is the true event type. ETA measures the correctness of the event semantic interpretation, ensuring that the model accurately identifies the event type. Correctly distinguishing between event types enhances downstream trading performance by enabling portfolio allocations tailored to different events.

To evaluate practical trading effectiveness, we report the \textbf{Sharpe Ratio (SR)}:
\[
\text{SR} = \frac{\mathbb{E}[r]}{\sigma_r}
\]
where $\mathbb{E}[r]$ is the expected return, and $\sigma_r$ is the standard deviation of returns. SR measures the risk-adjusted return, providing an indication of the strategy’s profitability relative to its volatility.

\textbf{Maximum Drawdown (MDD)}:
\[
\text{MDD} = \max_{t} \left( \frac{\text{peak}(t) - \text{trough}(t)}{\text{peak}(t)} \right)
\]
where $\text{peak}(t)$ is the highest portfolio value up to time $t$, and $\text{trough}(t)$ is the lowest portfolio value after time $t$. MDD quantifies the worst peak-to-trough loss, assessing the strategy's resilience during market downturns.

Together, these metrics provide a balanced evaluation of predictive reliability, decision consistency, and real-world trading performance.



\subsection{Event empirical study}

\subsubsection{Historical CAR Statistics}

\begin{figure}[t] 
    \centering
    \vspace{-6px}
    
    {\includegraphics[width=0.97\columnwidth]{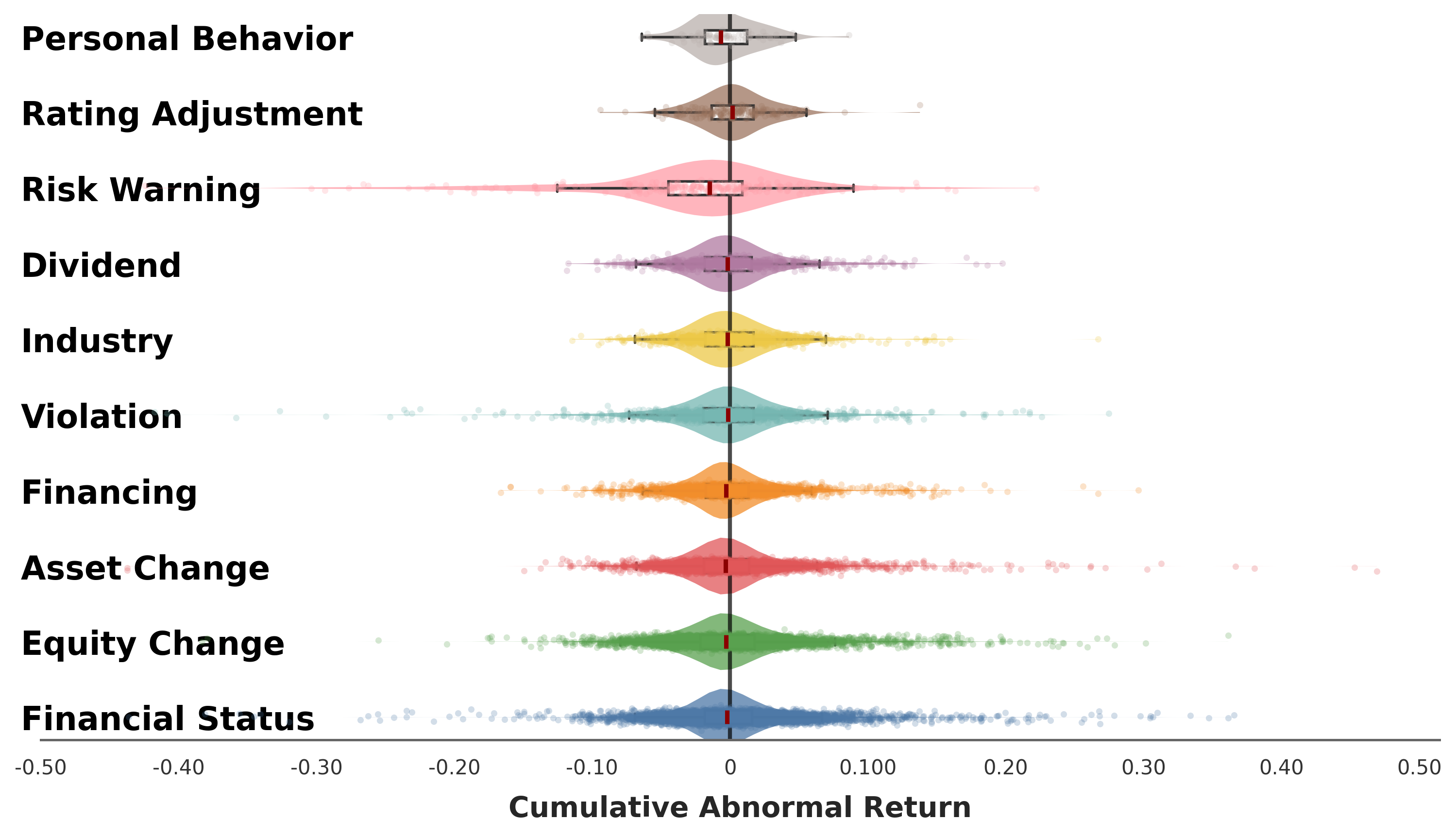}}
    \hfill
    \vspace{-8px}
    \caption{Distribution of post-event cumulative abnormal returns across event categories.}
    \vspace{-5px}
    \label{fig:distribution} 
\end{figure}

Figure~\ref{fig:distribution} illustrates the distribution of post-event cumulative abnormal returns 
across different event categories.
We observe substantial heterogeneity in both dispersion and tail behavior among event types.
Risk-related events such as \emph{Risk Warning} and \emph{Violation} exhibit wider distributions
and heavier tails, indicating higher uncertainty and asymmetric market reactions,
as well as a greater prevalence of extreme outcomes that may offer opportunities for outsized abnormal returns.
In contrast, routine corporate events including \emph{Dividend} and \emph{Industry} announcements
show more concentrated distributions around zero, suggesting relatively stable and predictable impacts.
These differences highlight that event categories are associated with distinct risk return profiles,
motivating event-aware modeling rather than uniform treatment of news signals.

\subsubsection{Historical Magnitude}

\begin{figure}[t] 
    \centering
    \vspace{-6px}
    
    {\includegraphics[width=0.97\columnwidth]{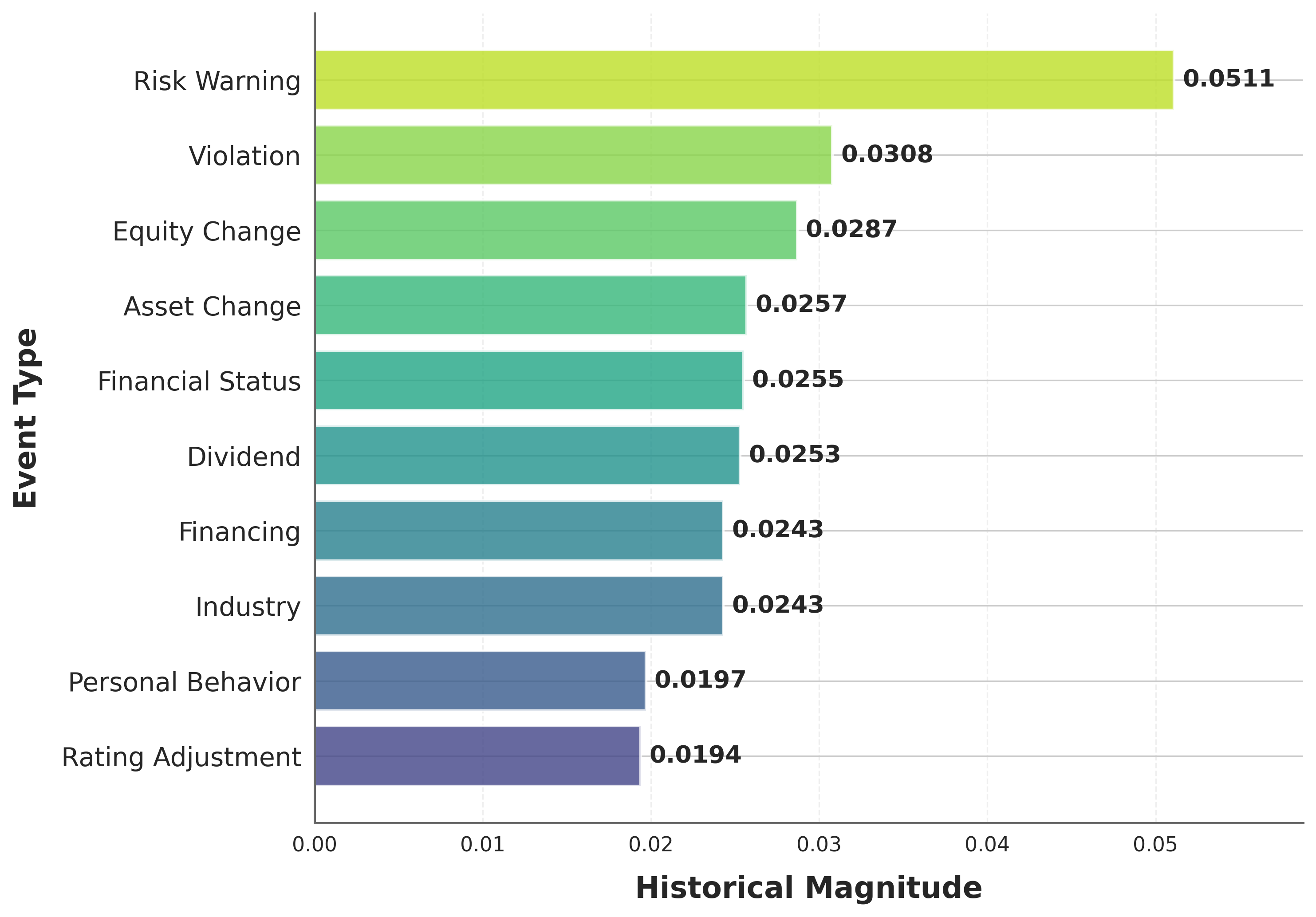}}
    \hfill
    \vspace{-8px}
    \caption{Mean absolute post-event cumulative abnormal returns by event category.}
    \vspace{-5px}
    \label{fig:event_weights_barplot} 
\end{figure}

Figure~\ref{fig:event_weights_barplot} reports the mean absolute post-event CAR for each event type,
reflecting the average strength of market reactions regardless of direction.
Events related to regulatory or risk disclosures, such as \emph{Risk Warning} and \emph{Violation},
exhibit the largest magnitudes, with mean absolute CAR exceeding 0.05 and 0.03, respectively,
indicating pronounced and economically meaningful market responses.
In contrast, softer informational events, including \emph{Personal Behavior} and \emph{Rating Adjustment},
show substantially smaller impacts, with average magnitudes below 0.02.
This ranking confirms that not all news carries equal economic significance
and suggests that event magnitude provides a natural prior for allocating trading attention and position size,
motivating magnitude-aware weighting in event-driven trading strategies.
To preserve temporal relevance, such magnitude-based weights are periodically re-estimated over rolling windows,
allowing the strategy to adapt to evolving market conditions and shifting event dynamics.

\subsubsection{Impact of Event Type Weighting on Trading Performance}

\begin{figure}[t] 
    \centering
    \vspace{-6px}
    
    {\includegraphics[width=0.97\columnwidth]{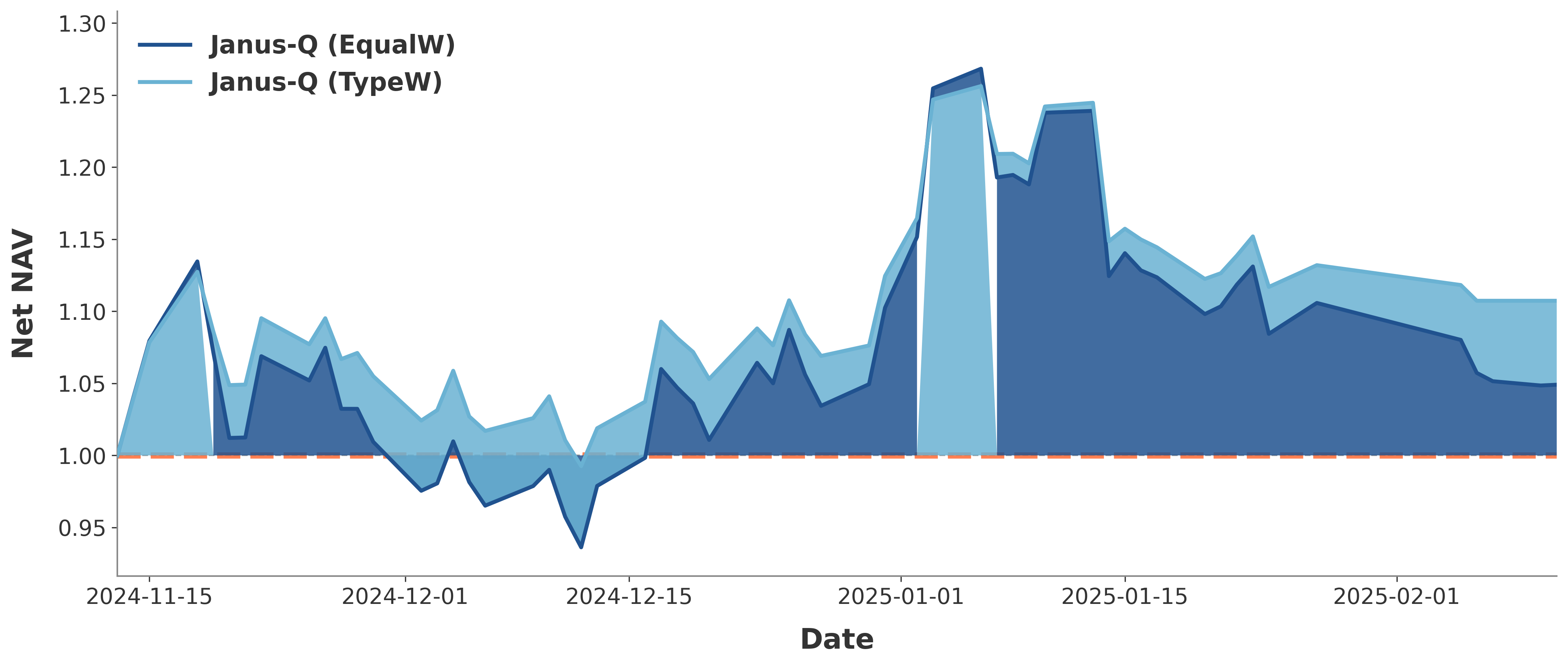}}
    \hfill
    \vspace{-8px}
    \caption{Net asset value (NAV) comparison between the Equal-Weighted and Type-Weighted Janus-Q strategies.}
    \vspace{-5px}
    \label{fig:two type weight} 
\end{figure}

Figure~\ref{fig:two type weight} compares the net asset value (NAV) trajectories of the Equal-Weighted and Type-Weighted Janus-Q strategies. While both strategies benefit from event-driven signals, the Type-Weighted variant achieves higher cumulative returns and exhibits greater stability during periods of heightened market volatility. Notably, between early and mid-December 2024, the Equal-Weighted strategy experiences pronounced drawdowns and temporarily enters negative territory, reflecting overexposure to low-impact or noisy events. In contrast, the Type-Weighted strategy remains comparatively stable throughout the same interval, as capital allocation is concentrated on historically high-impact events. The advantage becomes most apparent around high-impact event clusters, where weighting events by their historical CAR magnitudes enables more efficient capital allocation. These results demonstrate that incorporating event-level statistical priors effectively transforms heterogeneous market reactions into tangible trading gains.

\subsection{Risk Model Settings}
\label{app:risk_model}

To isolate event-driven abnormal returns from systematic risk exposures, we adopt a standard multi-factor risk model consistent with the \emph{CNE5} framework widely used in the Chinese equity market.\footnote{\url{https://www.msci.com/documents/10199/2935796a-0a80-4050-934a-12966d1e2518}}
CNE5 is a Barra-style equity risk model that decomposes stock returns into market, industry, and style-driven components, enabling robust neutralization of common risk premia.

In our implementation, the factor exposure vector $\mathbf{x}_{i,t}$ includes both industry factors and a set of non-financial style factors, covering:
\emph{Size}, \emph{Liquidity}, \emph{Volatility}, \emph{Momentum}, and \emph{Reversal}.
These factors capture systematic return patterns unrelated to firm-specific events, such as liquidity shocks, short-term reversals, or broad market sentiment.
We fully reproduce the factor construction and cross-sectional regression procedure following the CNE5 specification.
By removing the estimated factor-driven component from market-adjusted returns, the resulting abnormal returns more accurately reflect idiosyncratic, event-induced price movements rather than generic style or industry effects.

\subsection{Supplementary Experiments}

\subsubsection{Sensitive Analysis of Holding Period}


\begin{figure}[ht]
    \centering
    \vspace{-5px}

    \begin{subfigure}[t]{0.235\textwidth}
        \centering
        \includegraphics[width=\linewidth]{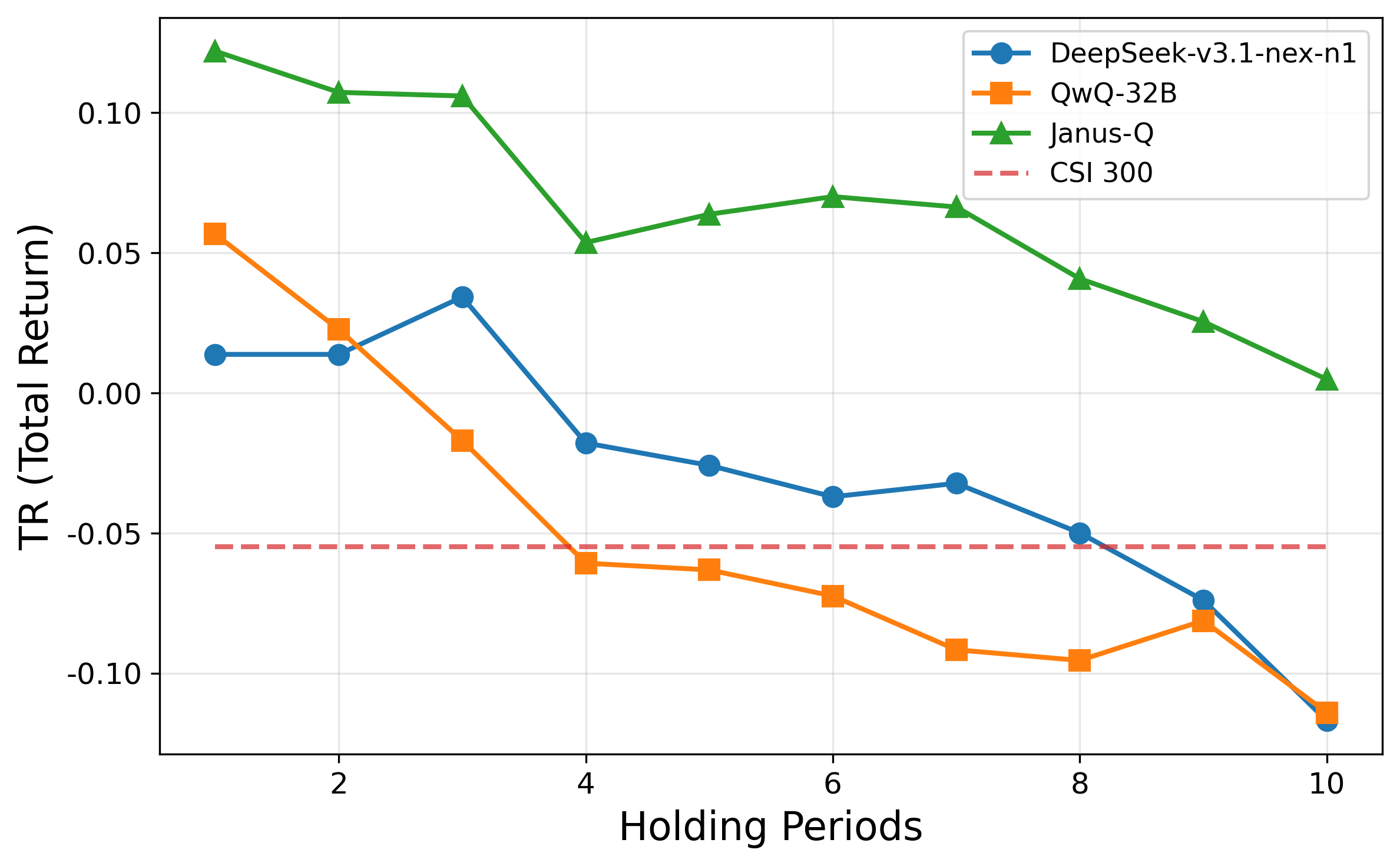}
        \caption{Total Return}
    \end{subfigure}
    \hfill
    \begin{subfigure}[t]{0.235\textwidth}
        \centering
        \includegraphics[width=\linewidth]{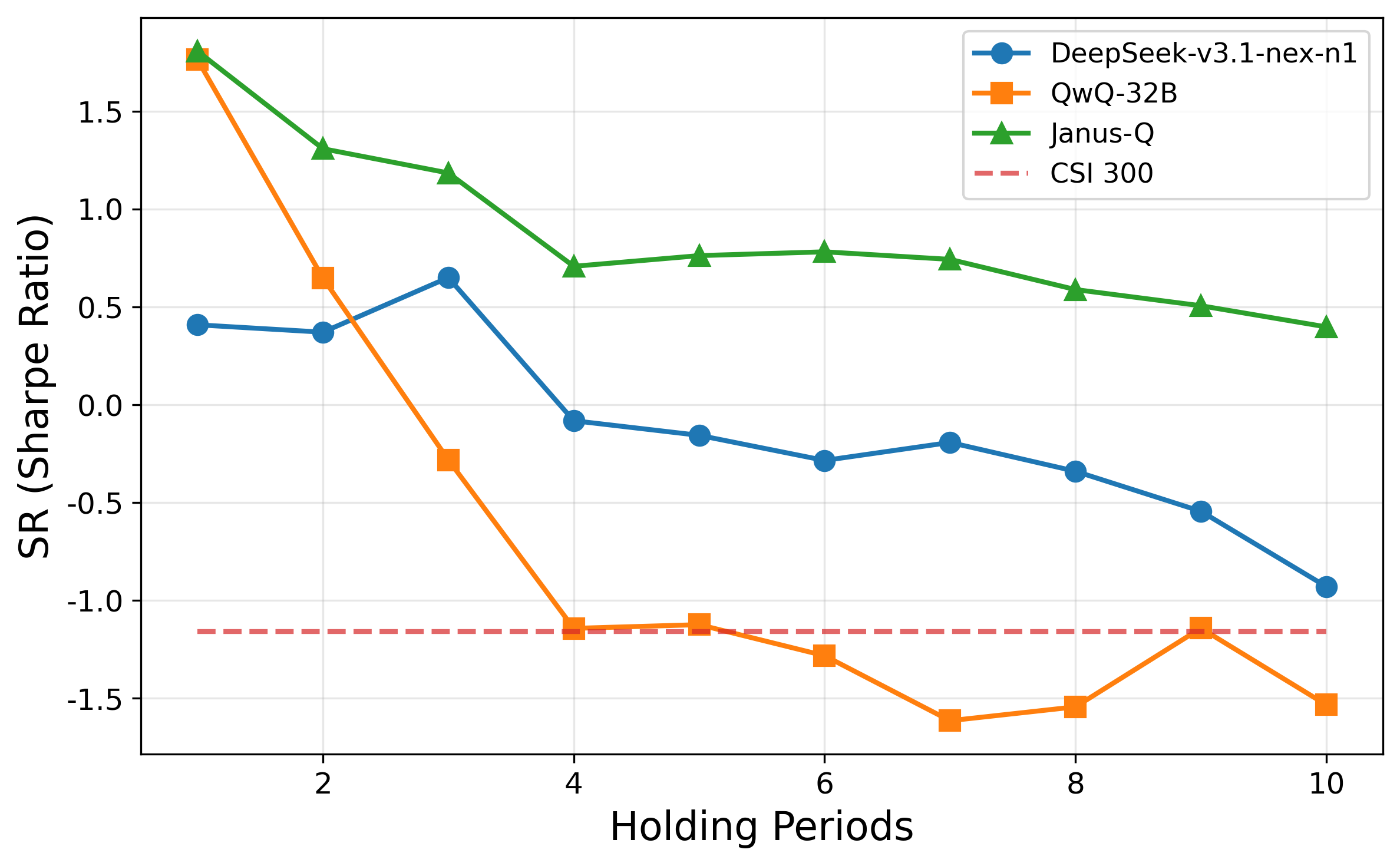}
        \caption{Sharpe Ratio}
    \end{subfigure}

    \vspace{-8px}
    \caption{Sensitivity analysis of backtesting performance with respect to the holding period.}
    \vspace{-5px}
    \label{fig:holding_period_sensitivity}
\end{figure}

Based on the experimental results in Table~\ref{tab:model_comparison}, we select the three best-performing models for further evaluation. Figure~\ref{fig:holding_period_sensitivity} examines the sensitivity of performance to the holding period. As the holding horizon increases, most baseline models exhibit pronounced performance degradation. For instance, QwQ-32B achieves a positive total return of 0.0567 and a Sharpe Ratio of 1.7642 at a one-day holding period but turns negative by day three, with its Sharpe Ratio falling below $-1.5$ at longer horizons. A similar though milder decline is observed for DeepSeek-v3.1-nex-n1, whose total return decreases from 0.0138 at one day to $-0.1169$ at ten days. This deterioration arises from the accumulation of overlapping event-driven positions, which increases exposure to unrelated market fluctuations as the holding horizon extends.

In contrast, Janus-Q demonstrates markedly greater robustness across holding periods. It reaches peak performance at short horizons, with a total return of 0.122 and a Sharpe Ratio of 1.8074 at one day, while maintaining positive returns up to nine days. Even at a ten-day horizon, its performance declines smoothly rather than collapsing, retaining a positive Sharpe Ratio of 0.398. This stability suggests that Janus-Q implicitly models the temporal decay of event influence and avoids excessive position persistence.

The CSI 300 benchmark remains unchanged across horizons, as it follows a static buy-and-hold strategy that is insensitive to event timing. Overall, these findings indicate that in unconstrained settings, event-driven strategies benefit from shorter holding periods that align with the transient nature of news impacts. Prolonged horizons lead to overlapping exposures and amplified drawdowns, while appropriate position control and horizon-aware exposure limits, examined later in our experiments, can help mitigate such degradation.

\begin{figure}[ht]
    \centering
    \vspace{-5px}

    \begin{subfigure}[t]{0.235\textwidth}
        \centering
        \includegraphics[width=\linewidth]{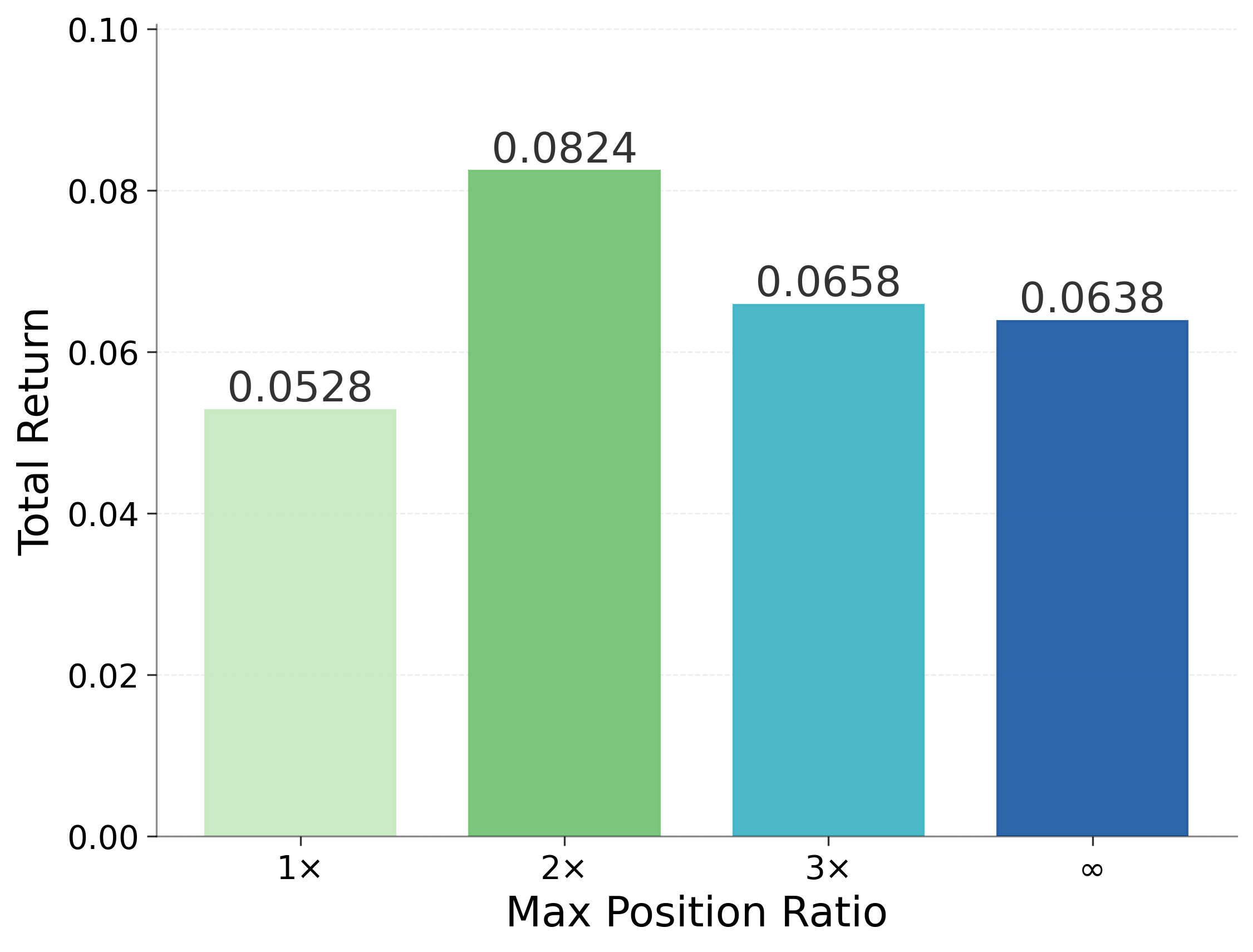}
        \caption{Holding Period = 5}
    \end{subfigure}
    \hfill
    \begin{subfigure}[t]{0.235\textwidth}
        \centering
        \includegraphics[width=\linewidth]{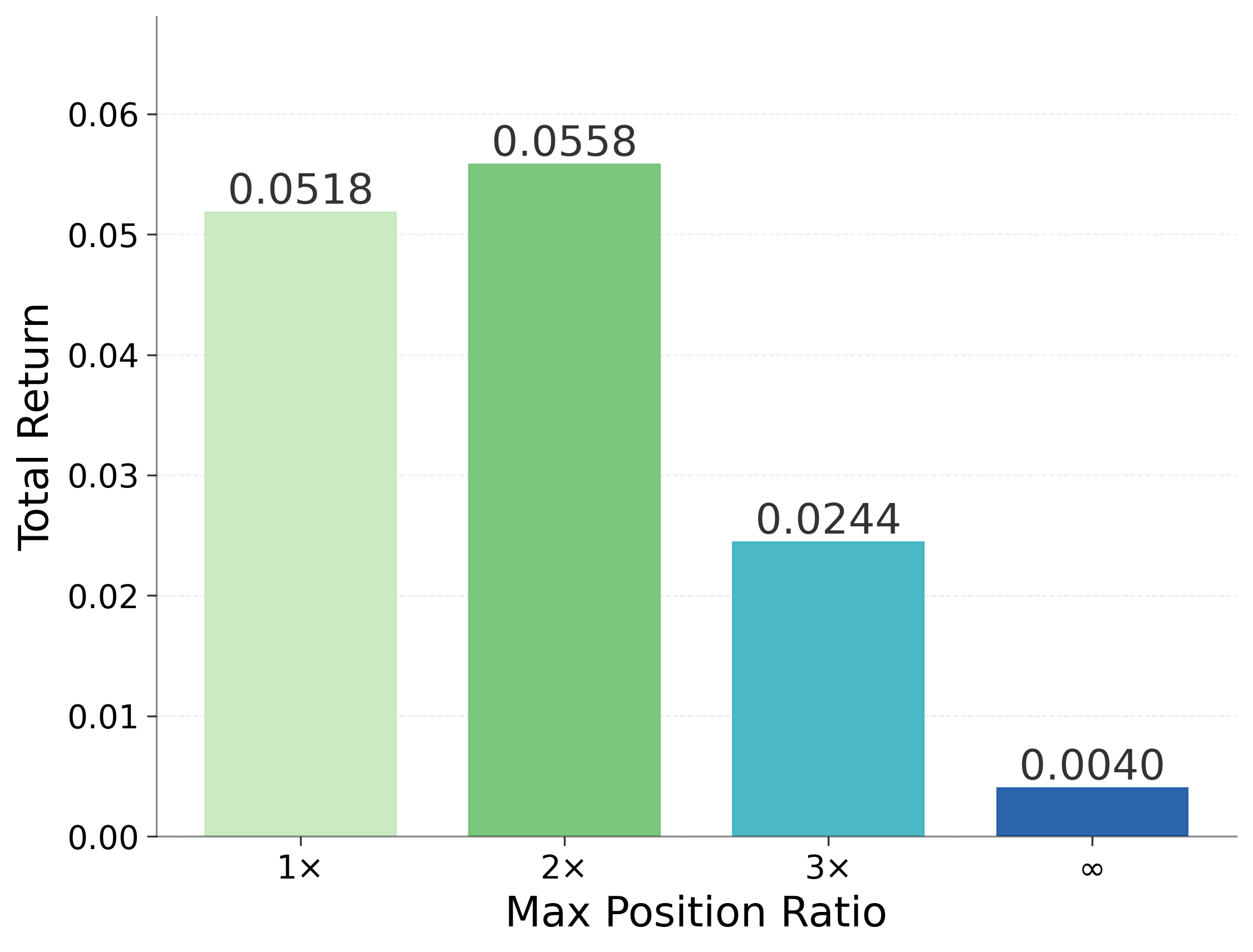}
        \caption{Holding Period = 10}
    \end{subfigure}

    \vspace{-8px}
    \caption{Sensitivity analysis of backtesting performance with respect to the maximum position ratio.}
    \vspace{-5px}
    \label{fig:max_position}
\end{figure}

\subsubsection{Sensitive Analysis of Maximum Position Ratio}

Building on the previous analysis of holding-period sensitivity, Figure~\ref{fig:max_position} presents a complementary experiment examining how position control affects performance under different trading frequencies. Specifically, it analyzes backtesting performance with respect to the maximum position ratio, using 5-day and 10-day holding periods as representative examples.

The maximum position ratio constrains the portfolio’s total notional exposure relative to its net asset value (NAV). A ratio of $k\times$ limits the combined exposure of all open positions to $k$ times the current NAV, thereby regulating both leverage and the extent to which multiple signals can be executed concurrently. The $\infty$ setting corresponds to the default configuration in the main experiments, where no explicit upper bound on the position ratio is applied.

As shown in Figure~\ref{fig:max_position}, excessively restrictive position limits (e.g., $1\times$) tend to impair performance, as the strategy becomes capital-constrained and must strictly adhere to the fixed holding period, missing profitable intermediate signals. Increasing the limit to moderate levels (e.g., $2\times$ or $3\times$) generally enhances performance by enabling partial overlap among positions and fuller utilization of concurrent signals. However, relaxing the constraint too much can again degrade performance, as excessive exposure leads to the accumulation of stale positions whose diminishing returns offset gains from newly arriving events.

Overall, these results reveal a trade-off between capital efficiency and signal freshness, indicating that moderate position limits provide the most balanced performance. Moreover, appropriate position control allows event-driven strategies to adapt to longer holding horizons and different trading frequencies, effectively stabilizing returns while maintaining responsiveness to new information.

\begin{algorithm}[t]
\caption{Hierarchical-Gated Reward Modeling (HGRM)}
\label{alg:hgrm}
\begin{algorithmic}[1]
\Require Model response $\mathcal{R}$, ground truth $(c,e)$, threshold $\tau$, transaction cost $\kappa$, clip bound $\rho$, discount $\alpha$
\Ensure Reward $R$

\State Parse prediction $\{\hat{c},\hat{d},\hat{s},\hat{e}\}$ from $\mathcal{R}$
\If{$\hat d$ is missing}
    \State $\hat d \gets \mathrm{sign}(\hat c)$
\EndIf
\If{$\hat s$ is missing}
    \State $\hat s \gets \texttt{strong}$ if $|\hat c|>\tau$ else $\texttt{weak}$
\EndIf

\State Derive $d \gets \mathrm{sign}(c)$,  $s \gets \texttt{strong}$ if $|c|>\tau$ else $\texttt{weak}$

\Statex \textit{// Hard gate: direction correctness}
\State Compute direction score $s_{\mathrm{dir}}(\hat d,d)$
\State Set hard gate $g_{\mathrm{dir}}\gets 1(s_{\mathrm{dir}}\ge 0)$

\Statex \textit{// Soft modulation: event-type consistency}
\State Compute event score $s_{\mathrm{evt}}(\hat e,e)$
\State Set discount factor $m_{\mathrm{evt}}\gets 1$ if $\hat e=e$ else $\alpha$

\Statex \textit{// Cost-aware trading payoff}
\State Compute $\mathrm{pnl}\gets \mathrm{PnL}(\hat d,c,\kappa)$

\If{$g_{\mathrm{dir}}=0$}
    \State $r_{\mathrm{pnl}}\gets 0$     \Comment{no trade executed}
\Else
    \State $r_{\mathrm{pnl}}\gets \mathrm{clip}(m_{\mathrm{evt}}\cdot \mathrm{Pnl},-\rho,\rho)$
\EndIf

\Statex \textit{// Magnitude shaping and process reward}
\If{$g_{\mathrm{dir}}=1$}
    \State $r_{\mathrm{mag}}\gets \exp(-|\hat c-c|/\sigma)$
    \State Compute process reward $r_{\mathrm{proc}}$
\Else
    \State $r_{\mathrm{mag}}\gets 0$; \ $r_{\mathrm{proc}}\gets 0$
\EndIf
\Statex \textit{// Final hierarchical reward}
\State $R\gets w_{\mathrm{dir}}s_{\mathrm{dir}}
+ g_{\mathrm{dir}}\Big(
    w_{\mathrm{evt}}s_{\mathrm{evt}}
    + w_{\mathrm{pnl}}r_{\mathrm{pnl}}
    + w_{\mathrm{mag}}r_{\mathrm{mag}}
    + w_{\mathrm{proc}}r_{\mathrm{proc}}
\Big)$
\State \Return $R$
\end{algorithmic}
\end{algorithm}

\subsection{Prompts in Janus-Q}

\subsubsection{Prompt for Reasoning Chain of Thought}

\begin{figure*}[t]
  \centering
  \includegraphics[
    width=0.92\textwidth,
    trim=40pt 5pt 40pt 20pt, 
    clip
  ]{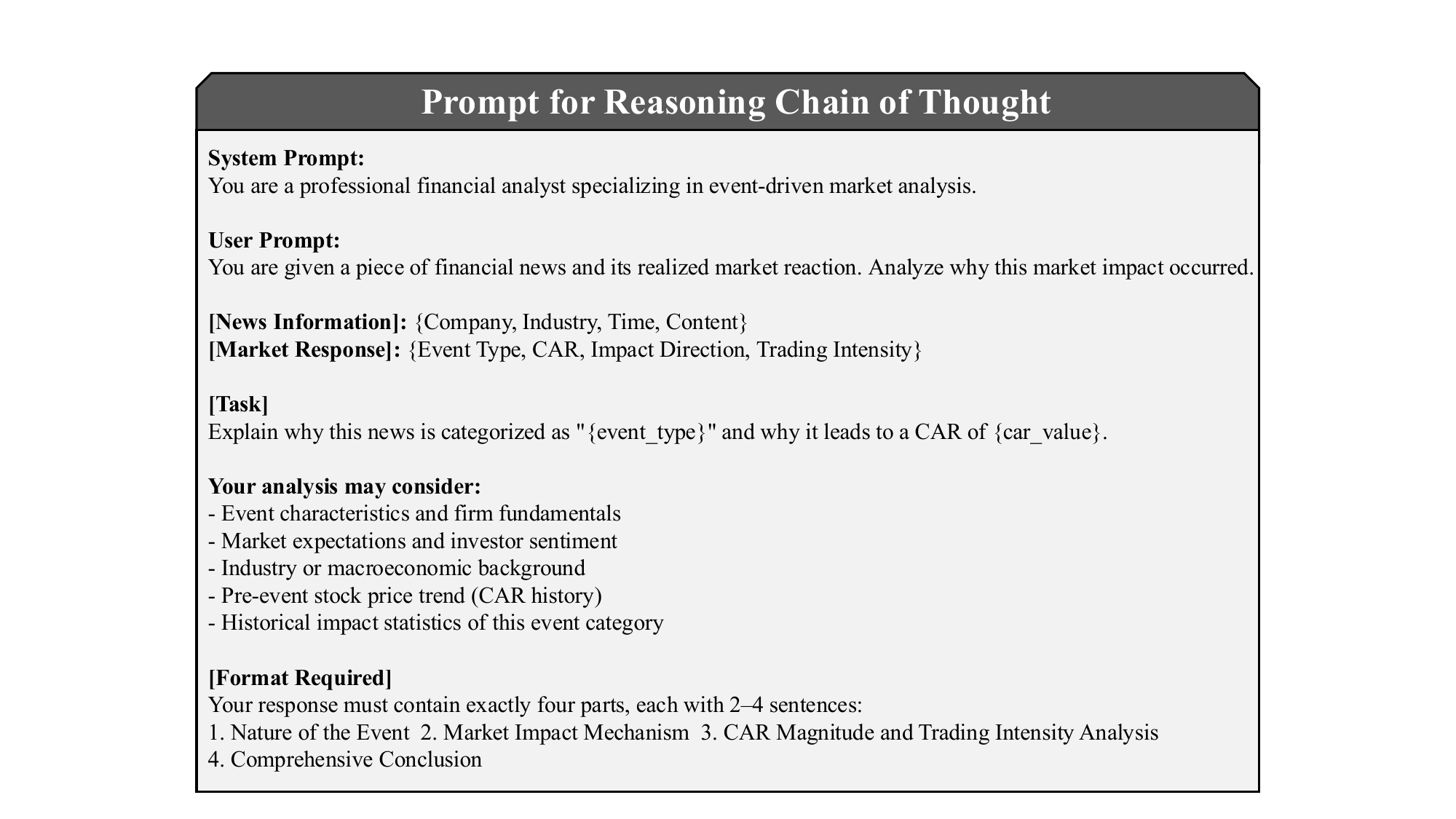} 
  \caption{Prompt for reasoning chain of thought.}
  \label{fig:cot_prompt}
\end{figure*}

Figure~\ref{fig:cot_prompt} illustrates the prompt used to elicit structured reasoning chains
for post-hoc analysis of event-driven market reactions.
The prompt guides the model to explain event categorization and observed cumulative abnormal returns
by explicitly considering event characteristics, market expectations, and historical context.
By enforcing a fixed multi-part structure, this prompt facilitates interpretable reasoning
and supports qualitative inspection of the model’s event-to-market understanding.

\subsubsection{Prompt for Training Template}

\begin{figure*}[t]
  \centering
  \includegraphics[
    width=0.92\textwidth,
    trim=40pt 5pt 40pt 20pt, 
    clip
  ]{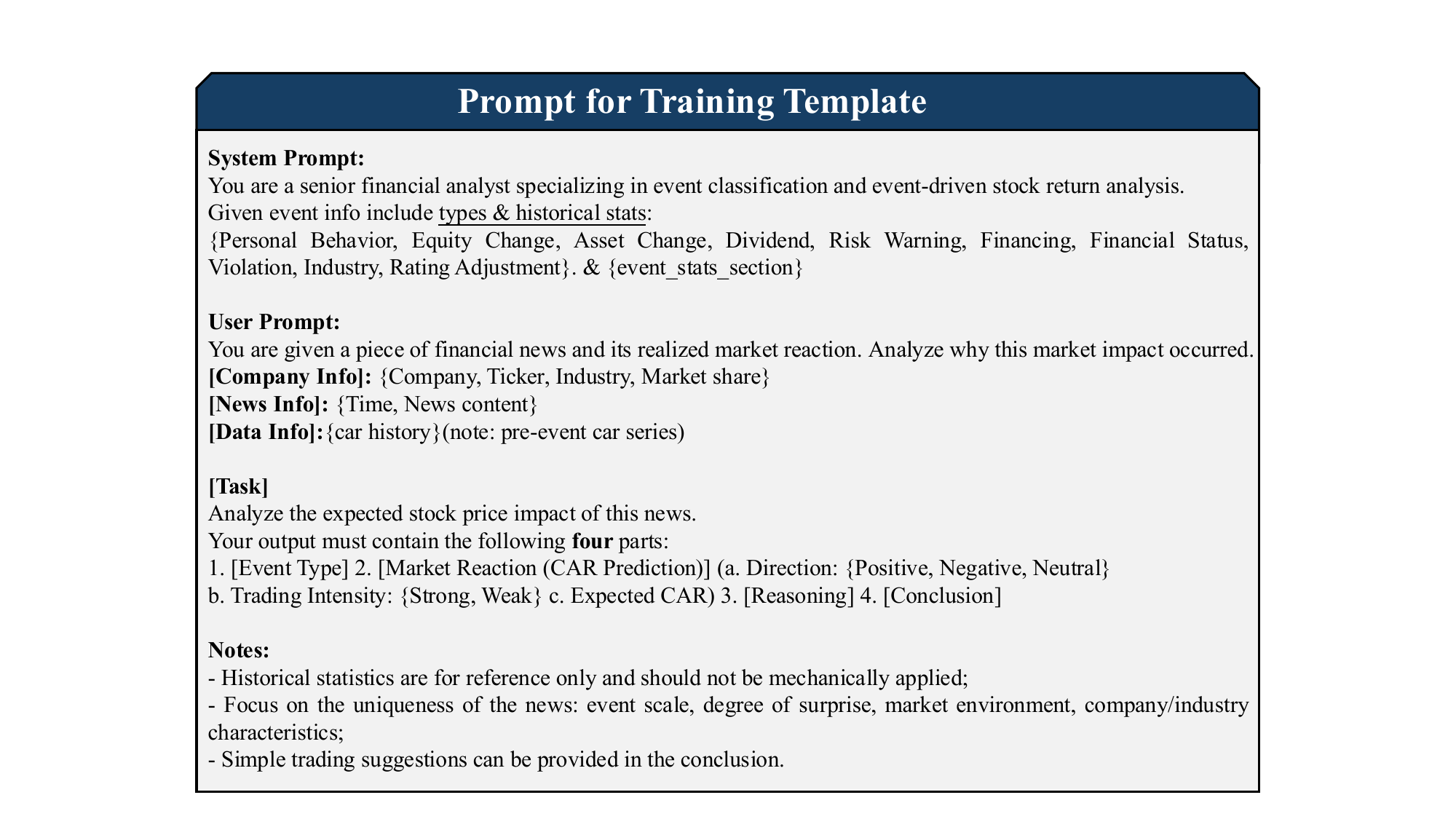} 
  \caption{Prompt for training template.}
  \label{fig:training_template}
\end{figure*}

Figure~\ref{fig:training_template} presents the training prompt template used during supervised fine-tuning.
The template requires the model to jointly predict event type, impact direction, trading intensity,
and expected abnormal return, while producing a concise rationale.
Historical CAR statistics are provided as contextual references rather than hard constraints,
encouraging the model to balance empirical priors with event-specific reasoning.
This design aligns model outputs with downstream decision objectives in event-driven trading.

\clearpage

\end{document}